\pdfoutput=1
\PassOptionsToPackage{table}{xcolor}
\documentclass{article}
\ifdefined\XeTeXversion\usepackage[T1]{fontenc}\fi
\usepackage{iclr2027_conference,times}

\usepackage{amsmath,amssymb,amsfonts}
\usepackage{etoolbox}    %
\usepackage{enumitem}
\usepackage{graphicx}
\usepackage{array}
\usepackage{booktabs}
\usepackage{tabularx}
\usepackage{xcolor}
\usepackage{multirow}
\usepackage{colortbl}    %
\usepackage{microtype}
\usepackage{subcaption}
\usepackage{wrapfig}     %
\usepackage{url}
\usepackage{hyperref}
\usepackage[capitalize,noabbrev]{cleveref}
\usepackage{algorithm}
\usepackage{algpseudocode}
\usepackage{tikz}
\usetikzlibrary{arrows.meta,positioning,fit,backgrounds,calc,shapes.geometric}
\usepackage{etoc}
\usepackage{listings}
\usepackage[framemethod=TikZ]{mdframed}

\definecolor{phcolor}{HTML}{B03A2E}   %
\definecolor{todocolor}{HTML}{7D3C98} %
\definecolor{harness}{HTML}{2A7F7F}
\definecolor{model}{HTML}{C8862A}
\definecolor{both}{HTML}{2F4B7C}
\definecolor{sftrow}{HTML}{EAF0FA}  %
\definecolor{rlrow}{HTML}{FBF1E4}   %
\definecolor{color5}{HTML}{006795}

\hypersetup{
  colorlinks = true,
  urlcolor   = blue,
  linkcolor  = color5,
  citecolor  = color5,
  pdftitle   = {CoTrace: Data Recipes for Training Terminal Agents with Harness-Model Co-Evolution},
  pdfauthor  = {Jixuan Chen, Jiaxin Zhang, Qinyuan Ye, Yada Pruksachatkun, Haoxiang Zhang, Jingming Zhuo, Yifan Zhang, Yutong Dai, Juntao Tan, Xiangyu Peng, Silvio Savarese, Zeyuan Chen, Lianhui Qin, Chien-Sheng Wu},
}

\definecolor{codegray}{rgb}{0.5,0.5,0.5}
\definecolor{codepurple}{rgb}{0.58,0,0.82}
\definecolor{backcolour}{rgb}{0.95,0.95,0.92}
\definecolor{PromptBg}{HTML}{EEF0FA}
\definecolor{PromptTitle}{HTML}{6F7DBF}
\definecolor{ConfigBg}{HTML}{EAF3F3}
\definecolor{ConfigTitle}{HTML}{2A7F7F}
\definecolor{BriefBg}{HTML}{FBF3E8}
\definecolor{BriefTitle}{HTML}{B07A2A}
\newmdenv[roundcorner=4pt,backgroundcolor=PromptBg,linecolor=PromptTitle,
  linewidth=0.6pt,innertopmargin=6pt,innerbottommargin=6pt,
  frametitlebackgroundcolor=PromptTitle,frametitlefont={\small\bfseries\color{white}},
  frametitleaboveskip=3pt,frametitlebelowskip=3pt,skipabove=6pt,skipbelow=6pt,nobreak=true]{promptbox}
\newmdenv[roundcorner=4pt,backgroundcolor=ConfigBg,linecolor=ConfigTitle,
  linewidth=0.6pt,innertopmargin=6pt,innerbottommargin=6pt,
  frametitlebackgroundcolor=ConfigTitle,frametitlefont={\small\bfseries\color{white}},
  frametitleaboveskip=3pt,frametitlebelowskip=3pt,skipabove=6pt,skipbelow=6pt,nobreak=true]{configbox}
\newmdenv[roundcorner=4pt,backgroundcolor=BriefBg,linecolor=BriefTitle,
  linewidth=0.6pt,innertopmargin=6pt,innerbottommargin=6pt,
  frametitlebackgroundcolor=BriefTitle,frametitlefont={\small\bfseries\color{white}},
  frametitleaboveskip=3pt,frametitlebelowskip=3pt,skipabove=6pt,skipbelow=6pt,nobreak=true]{briefbox}
\lstdefinestyle{plainbox}{language={},basicstyle=\ttfamily\scriptsize,breaklines=true,
  breakatwhitespace=false,columns=fullflexible,keepspaces=true,frame=none,numbers=none,
  backgroundcolor=\color{white!0},aboveskip=2pt,belowskip=2pt,showstringspaces=false,
  keywordstyle={},commentstyle={},stringstyle={},breakindent=1em,postbreak={}}

\crefname{appendix}{Appendix}{Appendices}\Crefname{appendix}{Appendix}{Appendices}

\newif\ifdraftnumbers
\draftnumbersfalse     %

\newcommand{\CONFIRMED}[1]{#1}

\newcommand{\methodname}{CoTrace}               %
\newcommand{\routername}{\textsc{FaultRoute}}   %

\newcommand{\NholdoutTmax}{\CONFIRMED{102}}
\newcommand{\NevolveTmax}{\CONFIRMED{50}}
\newcommand{\NpoolTmax}{\CONFIRMED{2{,}200}}
\newcommand{\NrlTmax}{\CONFIRMED{100}}

\newcommand{\ungatedStart}{\CONFIRMED{26.7}}
\newcommand{\ungatedEnd}{\CONFIRMED{0.0}}
\newcommand{\gatedBand}{\CONFIRMED{23--29}}

\newcommand{\routerConfThresh}{\CONFIRMED{0.6}}

\newcommand{\rlLR}{\CONFIRMED{$1\times 10^{-6}$}}
\newcommand{\rlGroup}{\CONFIRMED{$4\times 8$}}
\newcommand{\rlSubmitRateBefore}{\CONFIRMED{0}} %

\newcommand{\rlTruncBefore}{\CONFIRMED{0.88}}
\newcommand{\rlTruncAfter}{\CONFIRMED{0.00}}

\newcommand{\costEvolveHours}{\CONFIRMED{4.1}}

\newcommand{\tourN}{\CONFIRMED{5}}            %
\newcommand{\fanoutN}{\CONFIRMED{8}}          %
\newcommand{\fanoutKeep}{\CONFIRMED{2}}       %
\newcommand{\fanoutConc}{\CONFIRMED{4}}       %
\newcommand{\exploreEvery}{\CONFIRMED{3}}     %
\newcommand{\knnK}{\CONFIRMED{3}}             %
\newcommand{\probeSolved}{\CONFIRMED{2}}      %
\newcommand{\probeUnsolved}{\CONFIRMED{1}}    %
\newcommand{\probeMaxReg}{\CONFIRMED{1}}      %

\newcommand{\mixFail}{\CONFIRMED{10}}         %
\newcommand{\mixPass}{\CONFIRMED{6}}          %

\newcommand{\screenErrWarn}{\CONFIRMED{20}}   %

\newcommand{\routeBase}{\CONFIRMED{3.00}}       %
\newcommand{\routeMix}{\CONFIRMED{4.00}}        %
\newcommand{\routeTaxonomy}{\CONFIRMED{4.33}}   %
\newcommand{\routeCapability}{\CONFIRMED{6.00}} %
\newcommand{\routeN}{\CONFIRMED{15}}
\newcommand{\routeCorpus}{\CONFIRMED{400}}
\newcommand{\routeReps}{\CONFIRMED{3}}          %

\newcommand{\caseFourRone}{\CONFIRMED{0 $\rightarrow$ 2}}
\newcommand{\caseThirtyRtwo}{\CONFIRMED{0 $\rightarrow$ 3}}
\newcommand{\caseThirtyRthree}{\CONFIRMED{3 $\rightarrow$ 2}}
\newcommand{\caseThirtyRfour}{\CONFIRMED{2 $\rightarrow$ 4}}

\newcommand{\corpusMax}{\CONFIRMED{100}}       %

\newcommand{\evolveSetSize}{\CONFIRMED{50}}
\newcommand{\rotateFrom}{\CONFIRMED{2}}        %
\newcommand{\rotateMinSet}{\CONFIRMED{20}}     %

\newcommand{\innerTol}{\CONFIRMED{0.04}}        %

\newcommand{\probeNet}{\CONFIRMED{$g - 0.5r$}}  %
\newcommand{\refillSeed}{\CONFIRMED{42}}
\newcommand{\rotKept}{\CONFIRMED{14}}
\newcommand{\rotNew}{\CONFIRMED{36}}
\newcommand{\rotRetired}{\CONFIRMED{36}}
\newcommand{\rotNewR}{\CONFIRMED{0.40}}         %

\newcommand{\dHn}{\CONFIRMED{17}}            %
\newcommand{\dHchains}{\CONFIRMED{8}}        %
\newcommand{\dHpass}{\CONFIRMED{9}}          %

\newcommand{\dHmean}{\CONFIRMED{+0.18}}      %

\newcommand{\divTenEv}{\CONFIRMED{34 $\rightarrow$ 26 $\rightarrow$ 17}}
\newcommand{\divTenHo}{\CONFIRMED{75 $\rightarrow$ 81 $\rightarrow$ 83}}
\newcommand{\divTwelveEv}{\CONFIRMED{32 $\rightarrow$ 22 $\rightarrow$ 14 $\rightarrow$ 12}}
\newcommand{\divTwelveHo}{\CONFIRMED{80 $\rightarrow$ 80 $\rightarrow$ 80 $\rightarrow$ 83}}
\newcommand{\driftCorpusTasks}{\CONFIRMED{96}} %
\newcommand{\driftCorpusCurrent}{\CONFIRMED{14}} %
\newcommand{\driftEvolveCarried}{\CONFIRMED{32}} %

\newcommand{\anchorNineB}{\CONFIRMED{77--78}}  %
\newcommand{\anchorFourB}{\CONFIRMED{64--67}}  %

\newcommand{\frontHonly}{\CONFIRMED{86}}     %

\newcommand{\mixDeltas}{\CONFIRMED{$-2$, $-2$, $-4$}}

\newcommand{\nChangesets}{\CONFIRMED{216}}
\newcommand{\propAddProc}{\CONFIRMED{61}}\newcommand{\propPrompt}{\CONFIRMED{20}}
\newcommand{\propRemove}{\CONFIRMED{13}}\newcommand{\propNoop}{\CONFIRMED{25}}
\newcommand{\propNewTool}{\CONFIRMED{0}}
\newcommand{\addLoop}{\CONFIRMED{37}}\newcommand{\addDep}{\CONFIRMED{16}}
\newcommand{\addBudget}{\CONFIRMED{13}}\newcommand{\addOther}{\CONFIRMED{30}}
\newcommand{\nIncumbents}{\CONFIRMED{10}}\newcommand{\nAcceptedSlots}{\CONFIRMED{19}}
\newcommand{\accLoop}{\CONFIRMED{53}}\newcommand{\accBudget}{\CONFIRMED{26}}
\newcommand{\accDep}{\CONFIRMED{21}}\newcommand{\accPromptRewrite}{\CONFIRMED{3}}

\newcommand{\fourBanchor}{\CONFIRMED{64}}
\newcommand{\fourBnewAnchor}{\CONFIRMED{59}}
\newcommand{\twoBanchor}{\CONFIRMED{10}}
\newcommand{\twentySevenAnchor}{\CONFIRMED{97}}

\newcommand{\rlEpisodes}{\CONFIRMED{2{,}304}}
\newcommand{\rlTasks}{\CONFIRMED{100}}
\newcommand{\rlRespLen}{\CONFIRMED{16{,}384}}
\newcommand{\rlPerTurn}{\CONFIRMED{4{,}096}}
\newcommand{\rlMaxSteps}{\CONFIRMED{40}}
\newcommand{\rlPrompts}{\CONFIRMED{4}}
\newcommand{\rlSamples}{\CONFIRMED{8}}
\newcommand{\rlAsync}{\CONFIRMED{2}}
\newcommand{\rlLearners}{\CONFIRMED{6}}
\newcommand{\rlVllm}{\CONFIRMED{2}}
\newcommand{\rlGroupCap}{\CONFIRMED{8}}
\newcommand{\rlFrontierPct}{\CONFIRMED{20}}
\newcommand{\rlWall}{\CONFIRMED{7}}
\newcommand{\rlBeta}{\CONFIRMED{0.01}}\newcommand{\rlTV}{\CONFIRMED{0.1}}

\newcommand{\rlAnchorScore}{\CONFIRMED{78}}
\newcommand{\rlPeak}{\CONFIRMED{90}}
\newcommand{\rlFinalDrop}{\CONFIRMED{85}}
\newcommand{\rlHarnessShare}{\CONFIRMED{+5}}
\newcommand{\rlModelShare}{\CONFIRMED{+7}}

\newcommand{\flipHone}{\CONFIRMED{7 win / 7 loss}}
\newcommand{\flipRLone}{\CONFIRMED{8 win / 8 loss}}
\newcommand{\flipRLtwo}{\CONFIRMED{14 win / 9 loss}}
\newcommand{\flipRLthree}{\CONFIRMED{10 win / 8 loss}}
\newcommand{\flipHfour}{\CONFIRMED{3 win / 1 loss}}
\newcommand{\flipRLfour}{\CONFIRMED{4 win / 7 loss}}

\newcommand{\rlRetired}{\CONFIRMED{40}}   %

\newcommand{\sftGenTarget}{\CONFIRMED{100}}   %
\newcommand{\perTaskDefault}{\CONFIRMED{1}}   %
\newcommand{\histCap}{\CONFIRMED{40}}         %

\newcommand{\cost}[1]{\CONFIRMED{#1}}
\newcommand{\costHonly}{\cost{25}}     %
\newcommand{\costFixed}{\cost{77}}     %
\newcommand{\costRotating}{\cost{118}} %
\newcommand{\costMixed}{\cost{54}}     %
\newcommand{\costWinner}{\cost{47}}    %
\newcommand{\costRL}{\cost{62}}        %
\newcommand{\costGradShare}{\CONFIRMED{4}}   %
\newcommand{\costEvalShare}{\CONFIRMED{37--57}} %

\newcommand{\NtbTwoOne}{\CONFIRMED{89}}

\newcommand{\itersHonly}{\CONFIRMED{4}}
\newcommand{\itersFixed}{\PLACEHOLDER{2}}      %
\newcommand{\itersRotating}{\CONFIRMED{3}}
\newcommand{\itersMixed}{\CONFIRMED{3}}
\newcommand{\itersWinner}{\CONFIRMED{5}}
\newcommand{\itersRL}{\PLACEHOLDER{4}}         %
\newcommand{\totHonly}{\CONFIRMED{100}}

\newcommand{\gphHonly}{\CONFIRMED{25}}
\newcommand{\gphFixed}{\PLACEHOLDER{31}}
\newcommand{\gphRotating}{\CONFIRMED{51}}
\newcommand{\gphMixed}{\CONFIRMED{18}}
\newcommand{\gphWinner}{\CONFIRMED{24}}
\newcommand{\gphRL}{\PLACEHOLDER{21}}
\newcommand{\gphRLfinal}{\PLACEHOLDER{35}}
\newcommand{\tbBaseThree}{\needs}\newcommand{\tbSFTThree}{\needs}

\newcommand{\fourBRLInit}{\needs}\newcommand{\fourBRLPeak}{\needs}\newcommand{\fourBRLDelta}{\needs}
\newcommand{\fourBRLDH}{\needs}\newcommand{\fourBRLDM}{\needs}\newcommand{\costRLFourB}{\needs}

\newcommand{\tbBaseMean}{\CONFIRMED{18.7}}\newcommand{\tbBaseSd}{\CONFIRMED{1.5}}
\newcommand{\tbSFTMean}{\CONFIRMED{18.3}}\newcommand{\tbSFTSd}{\CONFIRMED{2.5}}

\renewcommand{\tbBaseThree}{\tbBaseMean{} $\pm$ \tbBaseSd{}}
\renewcommand{\tbSFTThree}{\tbSFTMean{} $\pm$ \tbSFTSd{}}

\newcommand{\NsweLite}{\CONFIRMED{300}}
\newcommand{\sweBaseRes}{\CONFIRMED{104}}\newcommand{\sweBaseResPct}{\CONFIRMED{34.7}}
\newcommand{\sweBaseFormat}{\CONFIRMED{55}}
\newcommand{\sweBaseCondPct}{\CONFIRMED{54.2}}
\newcommand{\sweSFTRes}{\CONFIRMED{74}}\newcommand{\sweSFTResPct}{\CONFIRMED{24.7}}
\newcommand{\sweSFTFormat}{\CONFIRMED{57}}
\newcommand{\sweSFTCondPct}{\CONFIRMED{51.0}}
\newcommand{\sweRLRes}{\CONFIRMED{110}}\newcommand{\sweRLResPct}{\CONFIRMED{36.7}}
\newcommand{\sweRLFormat}{\CONFIRMED{15}}
\newcommand{\sweRLCondPct}{\CONFIRMED{50.9}}
\newcommand{\sweBoth}{\CONFIRMED{77}}\newcommand{\sweBaseOnly}{\CONFIRMED{27}}\newcommand{\sweRLOnly}{\CONFIRMED{33}}
\newcommand{\sweRLGain}{\CONFIRMED{6}}

\newcommand{\fourBRLAnchor}{\CONFIRMED{67}}\newcommand{\fourBRLIone}{\CONFIRMED{74}}\newcommand{\fourBRLItwo}{\CONFIRMED{73}}
\newcommand{\fourBRLIthree}{\CONFIRMED{79}}
\renewcommand{\fourBRLInit}{67}\renewcommand{\fourBRLPeak}{79}\renewcommand{\fourBRLDelta}{+12}\renewcommand{\fourBRLDH}{0}\renewcommand{\fourBRLDM}{+12}
\renewcommand{\costRLFourB}{\CONFIRMED{100}}  %
\newcommand{\fourBRLEpisodesOne}{\CONFIRMED{5{,}120}}\newcommand{\fourBRLEpisodesLate}{\CONFIRMED{640}}
\newcommand{\fourBSFTAnchor}{\CONFIRMED{65}}\newcommand{\fourBSFTFinal}{\CONFIRMED{69}}
\newcommand{\fourBSFTDelta}{\CONFIRMED{+4}}\newcommand{\fourBSFTDH}{\CONFIRMED{+4}}
\newcommand{\fourBSFTDM}{\CONFIRMED{0}}
\newcommand{\fourBSFTModelSteps}{\CONFIRMED{$-5$, $-5$, $-6$}}
\newcommand{\fourBSFTHarnessSteps}{\CONFIRMED{$-2$, $+4$}}
\newcommand{\fourBSFTChains}{\CONFIRMED{five}}
\newcommand{\costSFTFourB}{\CONFIRMED{43}}
\newcommand{\fourBHoAnchor}{\CONFIRMED{65}}\newcommand{\fourBHoFinal}{\CONFIRMED{66}}
\newcommand{\fourBHoRounds}{\CONFIRMED{60, 60, 61, 62, 66}}

\newcommand{\costHonlyFourB}{\CONFIRMED{64}}   %
\renewcommand{\itersHonly}{\CONFIRMED{1}}\renewcommand{\totHonly}{\CONFIRMED{25}}\renewcommand{\gphHonly}{\CONFIRMED{6}}
\renewcommand{\itersFixed}{\CONFIRMED{3}}\renewcommand{\gphFixed}{\CONFIRMED{46}}
\renewcommand{\itersRL}{\CONFIRMED{4}}\renewcommand{\gphRL}{\CONFIRMED{21}}\renewcommand{\gphRLfinal}{\CONFIRMED{35}}
\newcommand{\reEvalFourBLo}{\CONFIRMED{61}}\newcommand{\reEvalFourBHi}{\CONFIRMED{66}}

\newcommand{\xcellOne}{\CONFIRMED{82}}
\newcommand{\sweOurBothBR}{92}\newcommand{\sweOurBaseOnlyR}{20}\newcommand{\sweOurRLOnlyB}{31}
\newcommand{\sweOurBaseOnlyS}{24}\newcommand{\sweOurSFTOnlyB}{17}
\newcommand{\sweOurSFTOnlyR}{18}\newcommand{\sweOurRLOnlyS}{36}

\newcommand{\sweOurBaseRes}{\CONFIRMED{112}}\newcommand{\sweOurBaseResPct}{\CONFIRMED{37.3}}\newcommand{\sweOurBaseEmpty}{\CONFIRMED{59}}\newcommand{\sweOurBaseUnres}{\CONFIRMED{129}}
\newcommand{\sweOurSFTRes}{\CONFIRMED{105}}\newcommand{\sweOurSFTResPct}{\CONFIRMED{35.0}}\newcommand{\sweOurSFTEmpty}{\CONFIRMED{64}}\newcommand{\sweOurSFTUnres}{\CONFIRMED{131}}
\newcommand{\sweOurRLRes}{\CONFIRMED{123}}\newcommand{\sweOurRLResPct}{\CONFIRMED{41.0}}\newcommand{\sweOurRLEmpty}{\CONFIRMED{36}}\newcommand{\sweOurRLUnres}{\CONFIRMED{141}}

\newcommand{\sweBaseNoPatch}{\CONFIRMED{108}}\newcommand{\sweSFTNoPatch}{\CONFIRMED{155}}\newcommand{\sweRLNoPatch}{\CONFIRMED{84}}
\newcommand{\sweOurBaseFinish}{143 / 147 / 10}\newcommand{\sweOurSFTFinish}{159 / 135 / 6}\newcommand{\sweOurRLFinish}{178 / 122 / 0}

\renewcommand{\arraystretch}{1.05}

\algrenewcommand\algorithmicindent{1.1em}
\algnewcommand{\LineComment}[1]{\State \(\triangleright\) #1}

\title{CoTrace: Data Recipes for Training Terminal Agents with Harness--Model Co-Evolution}

\author{%
\parbox[t]{\dimexpr\textwidth-2\tabcolsep\relax}{\centering
\bfseries
Jixuan~Chen$^{1,2}$\quad Jiaxin~Zhang$^{2}$\quad Qinyuan~Ye$^{2}$\quad
Yada~Pruksachatkun$^{2}$\quad Haoxiang~Zhang$^{1,2}$\quad Jingming~Zhuo$^{3}$\quad
Yifan~Zhang$^{2}$\quad Yutong~Dai$^{2}$\quad Juntao~Tan$^{2}$\quad Xiangyu~Peng$^{2}$\quad
Silvio~Savarese$^{2}$\quad Zeyuan~Chen$^{2}$\quad Lianhui~Qin$^{1}$\quad Chien-Sheng~Wu$^{2}$\\[3pt]
\mdseries
$^{1}$University of California, San Diego\quad
$^{2}$Salesforce AI Research\quad
$^{3}$University of Washington\\[2pt]
\small\ttfamily
jic182@ucsd.edu\\
\{jiaxin.zhang, qinyuan.ye, ypruksachatkun, wu.jason\}@salesforce.com
}}

\makeatletter
\pretocmd{\@maketitle}{\vspace*{-0.2in}}{}{\PackageWarning{main}{maketitle top patch failed}}
\patchcmd{\@maketitle}{{\LARGE\sc \@title\par}}{{\LARGE\sc \@title\par}\vskip 0.01in}{}{\PackageWarning{main}{maketitle title patch failed}}
\makeatother

\iclrfinalcopy

\begin{document}
\etocdepthtag.toc{maintext}
\maketitle
\lhead{Preprint.}

\begin{abstract}
Terminal-agent capability depends jointly on model weights and the runtime harness that formats prompts, binds tools, and handles error recovery. Existing harness--model co-evolution approaches improve both components, yet often treat trajectories produced during harness search as an undifferentiated replay buffer. This practice overlooks that a trajectory's value for model training depends on the harness under which it was generated. To systematically analyze this interface, we establish an alternating co-evolution framework that decouples harness search and policy training through component-wise promotion decisions. Within this framework, we introduce \methodname{}, a harness-aware data recipe that explicitly governs trajectory routing, provenance matching, and curriculum refresh. Under \methodname{}, recurring execution failures guide harness synthesis, while policy training is strictly conditioned on verified rollouts matched to the adopted runtime for supervised fine-tuning (SFT) or fresh online interactions for reinforcement learning (RL). On the Tmax promotion split, \methodname{} advances Qwen3.5-9B from 78 to 88 solved tasks under supervised fine-tuning while an online reinforcement variant reaches 90. Specifically, a compact harness-matched corpus produces steady model gains at substantially lower compute than much larger corpora pooled across sibling harnesses. Furthermore, evaluations on Terminal-Bench~2.1 and SWE-bench Lite show that out-of-distribution transfer depends fundamentally on harness compatibility, where maintaining consistency between training and evaluation runtimes prevents procedural execution breakdowns observed under foreign scaffolds.
\end{abstract}

\section{Introduction}
\label{sec:intro}

Recent advances have substantially improved autonomous coding and terminal agents across complex interactive environments \citep{swebench2024,terminalbench2026,tmax2026}. These agents combine two inseparable components: a language model that acts as the action-selection policy and a runtime harness that formats prompts, binds tools, manages context, and recovers from execution errors \citep{sweagent2024,openhands2024,harnessx2026}. Prior research has largely optimized either component in isolation, improving agent performance through verifier-grounded policy post-training under a fixed runtime \citep{tmax2026,swegym2024,swerl2025} or through automated harness search with frozen model weights \citep{ahe2026,harnessx2026,selfharness2026,metaharness2026}. Although recent systems interleave harness adaptation with policy training \citep{coharness2026,hase2026,evotrainer2026}, they leave the shared data interface between the two optimizers largely implicit. Because each update changes the conditions under which subsequent trajectories are collected, alternating the two processes without an explicit data recipe creates subtle failure modes. This motivates our central question: how should execution data flow between the policy and the harness during co-evolution?

\begin{figure}[t]
\centering
\resizebox{\linewidth}{!}{\input{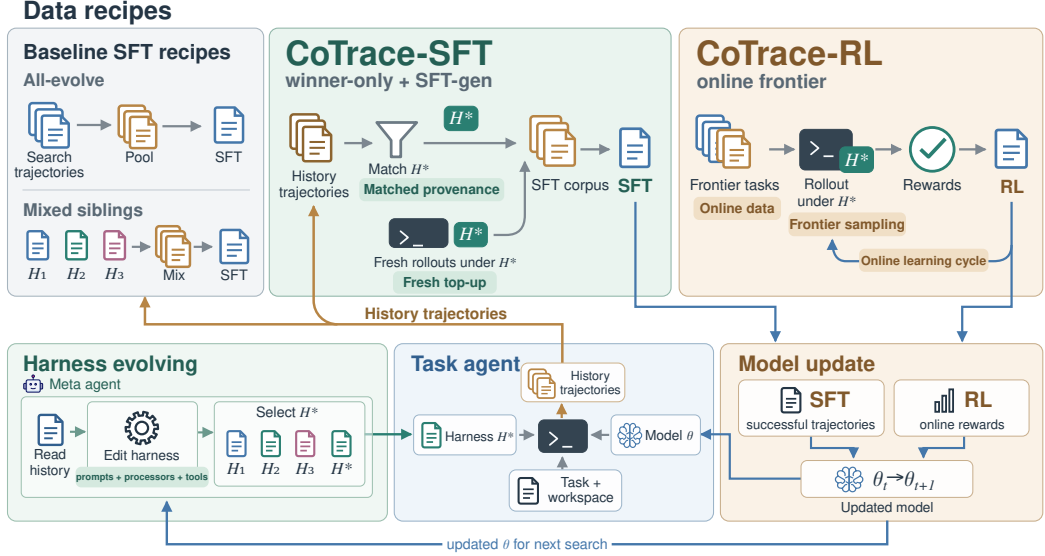}}
\caption{\textbf{Overview of \methodname{} and the shared data interface in model--harness co-evolution.}
\textbf{Bottom:} the closed loop. A task agent with the adopted harness $H^*$ and policy weights $\theta$ solves executable terminal tasks and fills a shared history of trajectories; harness evolution reads that history to edit prompts, processors and tools and selects the next $H^*$, while model training updates $\theta$ before the subsequent round of search.
\textbf{Top:} the data recipes that turn the history into model-training data, from the baseline supervised fine-tuning (SFT) recipes that pool every successful search trajectory (all-evolve) or successes across sibling candidate harnesses (mixed siblings), to \methodname{}-SFT, which keeps only trajectories whose provenance matches $H^*$ and tops them up with fresh rollouts under $H^*$, and \methodname{}-RL, which learns online by reinforcement learning (RL) from rewards on frontier tasks rolled out under $H^*$.}
\label{fig:overview}
\end{figure}

An execution trajectory has different value for the harness and the policy. Execution failures can expose missing recovery logic and thereby guide harness mutation, but they are poor targets for policy imitation. Conversely, successful executions may rely on processors specific to candidate harnesses that are later rejected; training on those trajectories can therefore mismatch the policy's data with the deployed runtime \citep{tajwar2024}. As the agent improves, successful rollouts also concentrate increasingly on solved tasks with diminishing learning value, while unresolved failures define a moving frontier of harder problems. Pooling all search trajectories can thus disconnect the two optimizers: in one of our experiments, harness updates add nine solved tasks, while the promotion procedure rejects every model update trained on pooled candidate successes.

We introduce \methodname{}, a harness-aware data recipe that explicitly governs how execution experience is filtered, matched, and refreshed across model--harness co-evolution. Operating over an alternating search and training pipeline, \methodname{} coordinates data flow through three targeted mechanisms. \textbf{Route} directs clustered execution failures to harness synthesis while restricting policy supervision to verified successes strictly matched to the adopted runtime, supplemented with fresh rollouts under that same harness. \textbf{Ratchet} attributes every accepted gain to one component by evaluating candidate harnesses under fixed weights and candidate policies under the adopted harness on a frozen promotion split. \textbf{Refresh} drives curriculum progression by retiring tasks only after they are both solved and incorporated into the training corpus, continuously redirecting exploration toward residual frontier tasks. On the 102-task Tmax split \citep{tmax2026}, our default supervised recipe advances Qwen3.5-9B from 78 to 88 solved tasks across alternating updates, while an online reinforcement variant reaches a peak of 90.

We evaluate these data recipes through targeted control experiments and cross-harness evaluations to determine when search experience translates into model learning. Tracking component-wise promotion decisions and data-routing strategies reveals that raw trajectory volume alone cannot guarantee policy improvement, whereas runtime compatibility proves decisive. While pooling 149--308 successful trajectories per iteration from exploratory sibling harnesses yields zero accepted model updates, our harness-matched recipe utilizes merely 30--50 trajectories to deliver consistent gains, achieving superior performance at the lowest per-iteration training cost of 47 GPU-hours. Cross-harness evaluations further demonstrate that learned capabilities remain tightly coupled to the runtime environment, consistently improving performance under the data-generating harness while degrading under mismatched alternatives. Crucially, external evaluations on Terminal-Bench~2.1 and SWE-bench Lite show that out-of-distribution (OOD) transfer is fundamentally governed by the co-evolved model--harness pair rather than the policy in isolation, where deploying the learned checkpoint within its co-evolved runtime dramatically suppresses early execution faults and no-patch failures, effectively unlocking the policy's underlying problem-solving capability across foreign domains without task-specific tuning.

In summary, our primary contributions are:
\begin{itemize}[leftmargin=*,topsep=2pt]
\item \textbf{An inspectable co-evolution framework} that records harness and model updates separately, with trajectory provenance and component-wise promotion decisions.
\item \textbf{A harness-aware data recipe}, combining failure routing, runtime-matched supervised trajectories, fresh generation, and curriculum refresh, evaluated against pooled search trajectories.
\item \textbf{An analysis of transfer across runtimes} showing where checkpoint-only gains fail to carry over and where runtime changes alter execution outcomes on external benchmarks.
\end{itemize}

\section{Method}
\label{sec:method}
\label{sec:system}
\label{sec:recipe}

We first formulate terminal-agent learning as alternating model--harness
optimization and then present \methodname{}, the data recipe that coordinates
the two update channels.

\subsection{Problem Formulation: Model--Harness Co-Evolution}
\label{sec:formulation}

An autonomous terminal agent is a pair of a parameterized policy $\theta$ and
an execution harness $H$. A task $x=(q_x,s_x,v_x)\sim\mathcal{T}$ specifies an
instruction $q_x$, an initial sandbox state $s_x$, and a deterministic verifier
$v_x$ of the terminal state; the harness formats prompt context, binds tools,
parses feedback, and handles error recovery, so executing the pair induces a
rollout distribution $\tau\sim\pi(\cdot\mid\theta,H,x)$ of interleaved tool
calls and observations. Model--harness co-evolution maximizes
\begin{equation}
    \max_{\theta, H} \; J(\theta, H) = \mathbb{E}_{x \sim \mathcal{T}, \, \tau \sim \pi(\cdot \mid \theta, H, x)} [v_x(\tau)],
    \label{eq:objective}
\end{equation}
which we optimize alternately over iterations $t$, $H_{t+1}\approx\arg\max_H J(\theta_t,H)$ and
$\theta_{t+1}\approx\arg\max_\theta J(\theta,H_{t+1})$. The two updates are
coupled through the traces $\tau$: each step produces the data the next
consumes, so co-evolution hinges on the recipe that routes failure evidence
to harness search and runtime-matched successes to the policy.

\subsection{\methodname{}: Data Recipe for Co-Evolution}
\label{sec:overview}

\methodname{} governs the data interface within an alternating co-evolution loop that coordinates harness search and model training over a versioned trajectory bank $\mathcal{B}_{0:t}$. Instead of pooling all generated traces into an undifferentiated replay buffer, \methodname{} synchronizes the information exchange through three coordinated operations. First, \textbf{Route} partitions execution traces into failure clusters that expose missing runtime recovery logic for harness synthesis, while filtering and matching verified successes to the adopted runtime for policy supervision. Second, \textbf{Ratchet} enforces coordinate promotion on the frozen split $\mathcal{V}$ with the complementary component fixed, guaranteeing that accepted gains are cleanly attributable and that regressions are discarded. Third, \textbf{Refresh} retires tasks that have provided verified training signal and replenishes the evolve set from an unvisited reservoir, ensuring that both search and training remain concentrated on the moving frontier. \Cref{alg:cotrace-simple} outlines this alternating workflow; see algorithm details in \Cref{app:algorithm}.

\begin{algorithm}[t]
\caption{\methodname{} Alternating Co-Evolution Loop (Single Iteration)}
\label{alg:cotrace-simple}
\small
\begin{algorithmic}[1]
\Require Incumbent pair $(\theta_t, H_t)$, evolve set $\mathcal{E}_t$, reservoir $\mathcal{T}_{\mathrm{pool}}$, frozen split $\mathcal{V}$, trajectory bank $\mathcal{B}_{0:t}$
\Ensure Updated pair $(\theta_{t+1}, H_{t+1})$, refreshed evolve set $\mathcal{E}_{t+1}$, updated bank $\mathcal{B}_{0:t+1}$

\vspace{0.4em}
\State \textbf{// Phase 1: Failure-Guided Harness Search and Attribution}
\State $\mathcal{D}^H_t \gets \textsc{RouteFailures}(\mathcal{B}_{0:t})$ \Comment{\textsc{Route}: cluster recurring runtime faults}
\State $H^{\mathrm{cand}} \gets \textsc{HarnessSearch}(H_t, \mathcal{D}^H_t, \mathcal{E}_t; \theta_t)$ \Comment{Explore runtime candidates with $\theta_t$ fixed}
\State $H_{t+1} \gets \textsc{RatchetEvaluation}(H^{\mathrm{cand}}, H_t \mid \theta_t, \mathcal{V})$ \Comment{\textsc{Ratchet}: adopt if $w_H > \ell_H$ on $\mathcal{V}$}

\vspace{0.4em}
\State \textbf{// Phase 2: Runtime-Matched Policy Learning and Attribution}
\If{Supervised Mode}
    \State $\mathcal{S}_t \gets \textsc{RouteMatchedSupervision}(\mathcal{B}_{0:t}, H_{t+1})$ \Comment{\textsc{Route}: match fingerprint $\phi(H_{t+1})$ and top up}
    \State $\theta' \gets \textsc{FineTunePolicy}(\theta_t, \mathcal{S}_t)$ \Comment{Update weights on matched demonstrations}
\Else
    \State $\theta' \gets \textsc{ReinforcePolicy}(\theta_t, H_{t+1}, \mathcal{T}_{\mathrm{pool}})$ \Comment{Sample rollouts and rewards online under $H_{t+1}$}
\EndIf
\State $\theta_{t+1} \gets \textsc{RatchetEvaluation}(\theta', \theta_t \mid H_{t+1}, \mathcal{V})$ \Comment{\textsc{Ratchet}: adopt if $w_M > \ell_M$ on $\mathcal{V}$}

\vspace{0.4em}
\State \textbf{// Phase 3: Moving Curriculum Frontier Progression}
\State $\mathcal{M}_t \gets \{x \in \mathcal{E}_t \mid \mathrm{solved}_t(x) \land \mathrm{harvested}_t(x)\}$ \Comment{Identify mastered tasks}
\State $\mathcal{E}_{t+1} \gets (\mathcal{E}_t \setminus \mathcal{M}_t) \cup \textsc{RefillDomainStratified}(\mathcal{T}_{\mathrm{pool}})$ \Comment{\textsc{Refresh}: advance active frontier}
\State $\mathcal{B}_{0:t+1} \gets \mathcal{B}_{0:t} \cup \textsc{HarvestIterationTraces}()$ \Comment{Log newly generated executions}

\vspace{0.3em}
\State \Return $(\theta_{t+1}, H_{t+1}), \mathcal{E}_{t+1}, \mathcal{B}_{0:t+1}$
\end{algorithmic}

\vspace{0.3em}
\hrule
\vspace{0.3em}
\scriptsize
\textbf{Notation and Operational Semantics}
\begin{itemize}[leftmargin=1.2em,itemsep=1pt,topsep=1pt,parsep=0pt]
    \item $w_H, \ell_H$ / $w_M, \ell_M$: number of tasks on $\mathcal{V}$ newly solved and newly broken by candidate $H^{\mathrm{cand}}$ under $\theta_t$, and candidate $\theta'$ under $H_{t+1}$
    \item $\phi(H_{t+1})$: hash signature capturing prompt templates, tool interface bindings, and observation processors of the adopted runtime
    \item $\mathrm{solved}_t(x)$: binary indicator that task $x$ is verified successful by the incumbent pair
    \item $\mathrm{harvested}_t(x)$: binary indicator that a verified execution trace for task $x$ is selected into the training corpus $\mathcal{S}_{0:t}$
\end{itemize}
\end{algorithm}

This alternating formulation abstracts away the underlying optimizers while isolating the data flow that couples them. In our implementation, harness search explores the typed configuration space of \textsc{HarnessX} \citep{harnessx2026} while model weights are optimized either by supervised fine-tuning with low-rank adaptation (LoRA) \citep{lora2022} or by outcome-driven reinforcement learning with the DPPO policy-gradient algorithm \citep{dppo2026,tmax2026,dapo2025}, with concrete search screens and training configurations documented in \Cref{sec:setup} and \Cref{app:method,app:rl}. The subsequent sections detail each of the three core operations (\textbf{Route}, \textbf{Ratchet}, and \textbf{Refresh}).
\label{sec:harness_search}
\label{sec:method:harness}
\label{sec:infra}

\subsubsection{Route: Separate Evidence for Each Optimizer}
\label{sec:route}
\label{sec:method:route}

A trajectory has different value for each optimizer, so \methodname{}
splits the bank into two views. The failure view
$\mathcal{D}^H_t=R_H(\mathcal{B}_{0:t})$ keeps executions with runtime faults
rather than environment or grader errors, clusters them by their earliest
unrecovered failure across tasks, and uses the resulting evidence to propose
harness changes (\Cref{app:routing}). The model view
$\mathcal{D}^M_t=R_M(\mathcal{B}_{0:t})$ keeps verified, well-formed
demonstrations outside $\mathcal{V}$, from which the SFT corpus is assembled by
\begin{equation}
    \max_{\mathcal{S}_t} \sum_{\tau \in \mathcal{S}_t} Q(\tau)
    \qquad \text{s.t.} \qquad
    |\mathcal{S}_t| \le C, \quad
    |\{\tau \in \mathcal{S}_t : \mathrm{task}(\tau) = x\}| \le c_r \;\; \forall x,
    \label{eq:corpus}
\end{equation}
where $Q$ scores trajectory quality, $C$ bounds the corpus, and $c_r$ caps
trajectories per task. A training-data recipe specifies five coupled choices:
trajectory \emph{provenance} (which harness generated it), \emph{per-task
sampling} ($c_r$), \emph{prompt and runtime conditioning} (whether the stored
prompt is that of the harness the policy will run under), \emph{fresh-rollout
generation}, and \emph{historical replay}. Provenance is recorded per
trajectory as a fingerprint $\phi(\tau)$ hashing the prompt template, tool
bindings, and processors it ran under, not merely a harness identifier.

\begin{itemize}[leftmargin=*,topsep=2pt]
  \item \textbf{\methodname{}-SFT} retains demonstrations whose fingerprint
  matches the adopted harness, prioritizes current successes over bounded
  historical replay, limits repeated examples from the same task, and adds
  fresh rollouts under the adopted harness when task coverage is insufficient.
  Thus every selected example reflects the runtime the updated policy will use.
  \item \textbf{\methodname{}-RL} uses fresh on-policy interactions and rewards
  collected under the adopted harness instead of constructing an offline
  corpus, so runtime alignment holds by construction.
\end{itemize}

\subsubsection{Ratchet: Isolate and Preserve Component Gains}
\label{sec:ratchet}
\label{sec:method:ratchet}

The evolve set $\mathcal{E}_t$ and the promotion split $\mathcal{V}$ serve distinct purposes, with $\mathcal{E}_t$ generating candidate updates and $\mathcal{V}$ determining whether each proposal replaces the incumbent. Every candidate $c$ is evaluated on $\mathcal{V}$ against the incumbent's task-by-task execution record. Letting $w$ and $\ell$ denote the number of tasks newly solved and newly broken on $\mathcal{V}$ respectively, a candidate is adopted if and only if
\begin{equation}
    \mathrm{Promote}(c) \iff w > \ell,
    \label{eq:ratchet}
\end{equation}
where exact ties are accepted only when resulting from a complete evaluation run free of container or infrastructure anomalies (\Cref{app:algorithm}). A candidate harness is scored with $\theta_t$ fixed, whereas a candidate policy is scored with $H_{t+1}$ fixed. This coordinate promotion protocol attributes each accepted increment to the isolated component that changed and prevents regressive updates from replacing the incumbent. When a model candidate is rejected, the framework gathers newly collected experience for the subsequent attempt rather than retraining on stale data.

\subsubsection{Refresh: Track the Learning Frontier}
\label{sec:refresh}
\label{sec:method:refresh}

At the end of each iteration, a task is retired from $\mathcal{E}_t$ once it
is both solved by the incumbent and \emph{harvested}, meaning selected into a
built SFT corpus $\mathcal{S}_{0:t}$; the evolve set is then replenished from
$\mathcal{T}_{\mathrm{pool}}$ (\Cref{eq:refresh}, \Cref{app:refresh}). Requiring
both conditions prevents a task from leaving before its success becomes usable
training data, while also requiring evidence that the incumbent has mastered
it. For online reinforcement, which has no persistent corpus, behavioral
success determines retirement.

\section{Experiments}
\label{sec:setup}

\subsection{Setup}
\label{sec:setup:setup}

\textbf{Tasks and splits.} Our primary testbed is built upon Tmax \citep{tmax2026}, a benchmark of \NpoolTmax{} executable terminal tasks equipped with containerized environments and programmatic verifiers. From this benchmark, we construct three disjoint splits with verified task-identifier separation (\Cref{app:tasks}), partitioning the pool into a held-out \emph{promotion} split $\mathcal{V}$ of \NholdoutTmax{} tasks dedicated to component adoption decisions, an active rotating \emph{evolve} set $\mathcal{E}_t$ of \NevolveTmax{} tasks for harness search and trajectory harvesting, and an unvisited reservoir $\mathcal{T}_{\mathrm{pool}}$ for curriculum replenishment alongside \NrlTmax{} tasks allocated for reinforcement learning. Out-of-distribution (OOD) generalization is evaluated on untouched external benchmarks that no stage of co-evolution optimizes against, spanning \NtbTwoOne{} tasks on Terminal-Bench~2.1 (TB2.1) \citep{terminalbench2026} and \NsweLite{} instances on SWE-bench Lite \citep{swebench2024} (\Cref{sec:rq:general}).

\textbf{Models and harnesses.} The task policy is instantiated with Qwen3.5-9B alongside Qwen3.5-4B as a supporting scale, guided by frontier meta-agents that propose runtime modifications in the typed configuration space of \textsc{HarnessX} \citep{harnessx2026}. Harness search explores this configuration space via a two-generation tournament ($\tourN{+}\tourN$), evaluating an initial pool of $\tourN$ proposals against the incumbent policy on the rotating evolve set before branching a second wave of $\tourN$ candidates from the top-performing variant, with full screening cascade details deferred to \Cref{app:screens}. For policy training, supervised updates use low-rank adaptation (LoRA) fine-tuning \citep{lora2022}, while reinforcement updates use DPPO \citep{dppo2026,tmax2026}, a PPO-style policy-gradient algorithm driven by binary outcome rewards, with group-relative advantage estimation, active sampling to discard zero-variance rollouts \citep{dapo2025}, and a binary total-variation trust region. Online reinforcement learning is scheduled once harness evolution establishes baseline competence to provide reliable reward signals, with complete search and training hyperparameters provided in \Cref{app:hparams,app:screens,app:rl}.

\textbf{Evaluation.} Performance is measured by the number of solved tasks on the frozen promotion split, where binary task verifiers govern promotion decisions relative to a strongly tuned baseline scaffold (\Cref{app:baseline}). At each update step, a candidate component is evaluated with the counterpart held at its best-so-far incumbent, ensuring that each evaluation isolates the marginal effect of a single component modification under deterministic greedy decoding. While repeated trials show modest run-to-run variation around two tasks (\Cref{app:noise}), our analyses focus on paired within-chain contrasts and consistent directional trends rather than isolated point estimates.

\subsection{Main Co-evolution Results}
\label{sec:res:main}

\Cref{tab:main} reports every tournament-search chain run to completion at
both scales, decomposed by the stage at which each update was installed
(harness candidates evaluated under the incumbent policy, model candidates
under the adopted harness).

\begin{table}[!htb]
\centering
\caption{\textbf{Harness--model co-evolution with different data recipes.}
We report the number of solved tasks on the frozen \NholdoutTmax-task promotion split of Tmax
\citep{tmax2026}; $\Delta_H$/$\Delta_M$ are cumulative harness and model
gains, \emph{Cost} is GPU-hours per iteration on one 8-GPU node. Shaded
rows are the \methodname{} recipes (blue supervised, orange reinforcement);
\textbf{bold}/\underline{underline} mark best and second-best per scale.
For non-tournament baseline comparisons using sequential harness search, see \Cref{tab:seqchains}.}
\label{tab:main}

\footnotesize
\setlength{\tabcolsep}{4.2pt}
\renewcommand{\arraystretch}{1.16}

\begin{tabular}{@{}lllcccccc@{}}
\toprule
\multicolumn{3}{c}{Setting}
  & \multicolumn{2}{c}{Score}
  & \multicolumn{3}{c}{Decomposition} & Cost \\
\cmidrule(lr){1-3}
\cmidrule(lr){4-5}
\cmidrule(lr){6-8}
\cmidrule(l){9-9}
Method & Harness search & Data recipe
  & Init & Evolved & $\Delta$ & $\Delta_H$ & $\Delta_M$
  & \shortstack{GPU-h\\ /\,iter} \\
\midrule

\rowcolor{black!6}
\multicolumn{9}{l}{\textit{Qwen3.5-9B-Thinking}} \\

Harness only\textsuperscript{a}
  & Tournament $5{+}5$
  & ---
  & 77 & 81
  & +4 & +4 & --- & \costHonly{} \\

Co-evolve w.\ SFT
  & Tournament $5{+}5$
  & Mixed siblings\textsuperscript{b}
  & 77 & 86
  & +9 & \textbf{+9} & 0 & \costMixed{} \\

\rowcolor{sftrow}
\methodname{}-SFT
  & Tournament $5{+}5$
  & Winner-only + SFT-gen\textsuperscript{c}
  & 78 & \underline{88}
  & \underline{+10} & +4 & \underline{+6} & \costWinner{} \\

\rowcolor{rlrow}
\methodname{}-RL\textsuperscript{d}
  & Tournament $5{+}5$
  & Winner-only harness
  & 78 & \textbf{90}
  & \textbf{+12} & \underline{+5} & \textbf{+7} & \costRL{} \\

\midrule
\rowcolor{black!6}
\multicolumn{9}{l}{\textit{Qwen3.5-4B-Thinking}} \\

Harness only
  & Tournament $5{+}5$
  & ---
  & \fourBHoAnchor{} & \fourBHoFinal{}
  & +1 & \underline{+1} & --- & \costHonlyFourB{} \\

\rowcolor{sftrow}
\methodname{}-SFT
  & Tournament $5{+}5$
  & Winner-only + SFT-gen\textsuperscript{c}
  & \fourBSFTAnchor{} & \underline{\fourBSFTFinal{}}
  & \underline{\fourBSFTDelta{}} & \textbf{\fourBSFTDH{}} & \fourBSFTDM{}
  & \costSFTFourB{} \\

\rowcolor{rlrow}
\methodname{}-RL\textsuperscript{d}
  & Tournament $5{+}5$
  & Winner-only harness
  & \fourBRLInit{} & \textbf{\fourBRLPeak{}}
  & \textbf{\fourBRLDelta{}} & \fourBRLDH{} & \textbf{\fourBRLDM{}} & \costRLFourB{} \\

\bottomrule
\end{tabular}
\vspace{2pt}
\begin{minipage}{\linewidth}\scriptsize
\textsuperscript{a}\ Evaluated using each round's top harness candidate for 9B, and the promoted incumbent after five rounds for 4B.
\textsuperscript{b}\ Pools verified successes across all candidate harnesses evaluated during tournament search under their original prompts.
\textsuperscript{c}\ Restricts offline supervision to trajectories matching the adopted harness fingerprint $\phi(H_{t+1})$, topped up with fresh rollouts under $H_{t+1}$.
\textsuperscript{d}\ Samples rollouts and rewards online under the adopted harness $H_{t+1}$; \emph{Evolved} denotes the final promoted incumbent (\Cref{app:chains}).
\end{minipage}
\end{table}

\begin{figure}[!htb]
\centering
\includegraphics[width=\linewidth]{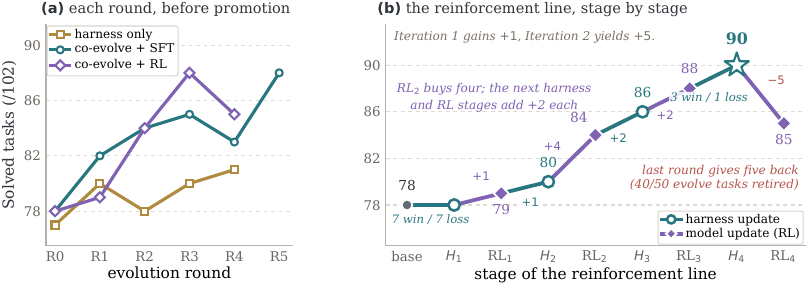}
\caption{\textbf{Dynamics of model--harness co-evolution.} \textbf{(a)} Each
round's artifact before promotion (harness-only: the round's best harness candidate; co-evolution lines: that round's candidate model checkpoint). \textbf{(b)} The 9B
reinforcement chain stage by stage; segment labels are the net task change or
the paired win/loss record where recorded (\Cref{tab:flips}). The incumbent
reaches \rlPeak{}; the next reinforcement candidate scores \rlFinalDrop{} and
is rejected.}
\label{fig:trajectories}
\end{figure}

On Qwen3.5-9B, the harness-only control reaches 81 from 77. Under tournament
search, mixed-siblings co-evolution reaches 86, with all nine accepted tasks
attributed to harness updates. \methodname{}-SFT reaches 88 from 78, with
cumulative gains of $+4$ from harness updates and $+6$ from model updates.
\methodname{}-RL reaches an incumbent of \rlPeak{}, with \rlHarnessShare{}
attributed to harness updates and \rlModelShare{} to reinforcement updates;
the reinforcement candidate that followed scores \rlFinalDrop{} and is
rejected. The two-task gap between \methodname{}-SFT and mixed siblings
lies within the measured evaluation variation (\Cref{app:noise}); the
contrast that matters is therefore the $+6$ against $0$ through the model
channel. \Cref{fig:trajectories} shows how these totals
accumulated: the score each round produced before promotion and the reinforcement
chain stage by stage.

At 4B, the contribution pattern changes. \methodname{}-SFT reaches
\fourBSFTFinal{} from \fourBSFTAnchor{} entirely through the harness channel,
with all three of its model updates rejected. By contrast, the reinforcement
chain moves from \fourBRLInit{} to \fourBRLPeak{} entirely through model
updates, with every harness candidate rejected; the frozen-weight control ends
one task above its anchor. We therefore use these results to study not
only whether co-evolution improves the pair, but which channel contributes
under different training regimes and policy scales (\Cref{sec:analysis}).

\subsection{Data Recipes and Model-Side Updates}
\label{sec:res:recipe}

\Cref{tab:recipe} summarizes what each recipe provides to the trainer.
Under the controlled tournament-search comparison at 9B, mixed siblings
supplies 149--308 trajectories per iteration but yields no accepted model
update, whereas winner-only $+$ SFT-gen uses 30--50 and contributes two
accepted updates totaling $+6$ at \costWinner{} GPU-hours per iteration
against \costMixed{}. At 4B neither supervised construction yields model-side
gain, while online reinforcement does.

Two features of \Cref{tab:recipe} are important for interpretation. First, the recipes
differ jointly in harness provenance, prompt conditioning, per-task cap,
coverage, freshness and historical replay, so the table compares complete
recipes, and \Cref{sec:rq:recipe} asks which of these choices the record can
attribute the difference to. Second, a trajectory is not one optimization
example: one execution contributes several prompt/completion pairs, and the
pair counts overlap, 618--1335 per iteration for mixed siblings against
526--948 for winner-only $+$ SFT-gen (\Cref{tab:manifests}), so the five-fold
difference in trajectories corresponds to comparable numbers of training
examples and comparable gradient compute. Per-iteration cost likewise differs
from total cost because chain lengths differ: with the rounded values of
\Cref{tab:main}, mixed siblings runs for about $3\times54=162$ GPU-hours and
\methodname{}-SFT for $5\times47=235$.

\begin{table}[!htb]
\centering
\caption{\textbf{Training data from each mixture and the resulting promotion
decisions.} Trajectories and unique tasks per iteration (\Cref{tab:manifests}),
the construction choices of \Cref{sec:method:route}, and accepted model
updates out of those attempted with their total $\Sigma\Delta_M$.}
\label{tab:recipe}
\label{tab:factors}
\scriptsize
\setlength{\tabcolsep}{3pt}\renewcommand{\arraystretch}{1.15}
\begin{tabular}{@{}lcccccccc@{}}
\toprule
Mixture & \shortstack{Traj.\ (tasks)\\per iter.} & \shortstack{Per-task\\cap} & \shortstack{Harness-\\matched} & \shortstack{Prompt\\conditioned} & \shortstack{Fresh\\top-up} & \shortstack{History\\cap} & \shortstack{9B: accepted,\\$\Sigma\Delta_M$} & \shortstack{4B:\\$\Sigma\Delta_M$} \\
\midrule
Mixed siblings               & 149--308 (38--88) & 8 & $\times$ & $\times$ & $\times$ & --- & 0/3, 0 & $0$ \\
\rowcolor{sftrow}
\methodname{}-SFT: winner-only $+$ SFT-gen & 30--50 (30--50) & 1 & \checkmark & \checkmark & \checkmark & 40\% & 2/5, $\mathbf{+6}$ & $0$ \\
\rowcolor{rlrow}
\methodname{}-RL: online frontier & \rlEpisodes{} ep. & --- & on-policy & --- & on-policy & --- & 3/4, $\mathbf{+7}$ & $\mathbf{+12}$ \\
\bottomrule
\end{tabular}
\end{table}

\section{Analysis}
\label{sec:analysis}

\subsection{RQ1: How do the two optimization channels contribute over a co-evolution chain?}
\label{sec:rq:relay}
\label{sec:res:relay}
\label{sec:rq:drift}
\label{sec:res:mismatch}
\label{sec:rq:cost}

\begin{figure}[!htb]
\centering
\includegraphics[width=\linewidth]{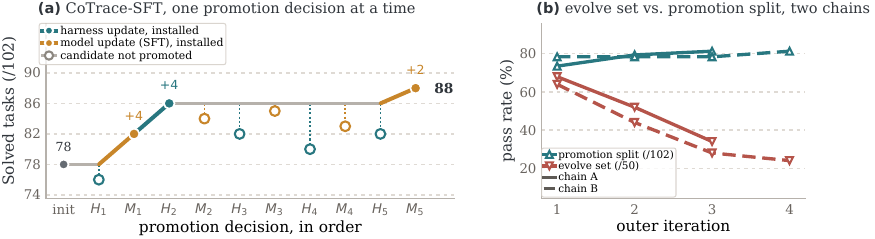}
\caption{\textbf{Promotion decisions and the moving curriculum.} \textbf{(a)}
\methodname{}-SFT one promotion decision at a time: filled markers are installed
updates (teal harness, amber model), hollow markers rejected candidates.
\textbf{(b)} Two chains: the evolve-set score falls as solved tasks retire
while the promotion score rises.}
\label{fig:rq1}
\end{figure}

\textbf{Accepted gains arise from both channels at different stages of the
chain.} \Cref{fig:trajectories}(b) and \Cref{fig:rq1}(a) open the two
\methodname{} chains at the level of individual promotion decisions. In the
supervised chain a model update ($+4$) precedes the first productive harness
search ($+4$ under the updated policy), six candidates are then rejected and
a late update adds $+2$; the reinforcement chain alternates the same way,
reaching \rlPeak{} before its last candidate is rejected at \rlFinalDrop{}. A pair-level total therefore conceals which optimizer is
active at each stage, and the channel a search finds productive depends on
the policy it runs against; the promotion rule filters a noisy stream, since
only \dHpass{} of \dHn{} search-winning harness candidates improved the
promotion split (\Cref{app:traces}).

\textbf{Cross-evaluation confirms independent component gains.}
Component-wise promotion attributes each accepted update while holding the
other component fixed.\hfill\null
\begin{wraptable}{r}{0.38\linewidth}
\centering
\captionsetup{font=small,justification=raggedright,singlelinecheck=false,skip=3pt}
\caption{\textbf{Completed cross-evaluation.}
Promotion-split scores for \methodname{}-SFT, iterations 1--2.}
\label{tab:cross}
\footnotesize
\setlength{\tabcolsep}{3pt}\renewcommand{\arraystretch}{1.12}
\begin{tabular}{@{}lcc@{}}
\toprule
Harness & $\theta_0$ (base) & $\theta_1$ (installed) \\
\midrule
previous $H_t$ & 78 & 82 \\
adopted $H_{t+1}$ & \xcellOne{} & \textbf{86} \\
\bottomrule
\end{tabular}
\end{wraptable}
The $2{\times}2$ cross-evaluation in \Cref{tab:cross}
then identifies the model gain under each harness,
$\Delta_M(H)=J(\theta_{t+1},H)-J(\theta_t,H)$, and the difference-in-differences
$I_t=\Delta_M(H_{t+1})-\Delta_M(H_t)$ measures how strongly a model update's
effect depends on the harness. In the \methodname{}-SFT chain, the harness adopted after
the model update is worth $+4$ under the base policy as well
($J(\theta_0,H_2)=\xcellOne{}$), so
$\Delta_M(H_0)=\Delta_M(H_2)=+4$ and $I=0$; for that transition the gains add
rather than interact. Together with the alternating promotion record, this
result shows both channels contributing within the same chain and adding
cleanly in the completed transition.

\textbf{As the pair improves, the optimization distribution moves toward the
residual frontier.} Solved-and-harvested tasks retire and unsolved ones stay,
so the evolve set hardens while the promotion score rises (\Cref{fig:rq1}b):
\divTenEv{} of \evolveSetSize{} against \divTenHo{} of \NholdoutTmax{} in one
chain. The record of the two \methodname{} chains shows the consequence. In the supervised chain every
harness candidate after the second iteration scores below the incumbent
($-4$, $-6$ and $-4$; \Cref{tab:ledger}), and in the reinforcement chain the
final candidate, trained after \rlRetired{} of \evolveSetSize{} evolve tasks had
retired, is the only stage that gives tasks back (\rlFinalDrop{} against the
incumbent \rlPeak{}). Late candidates are proposed and trained on a residual
set in which successes are scarce, and we read this pattern as the current
curriculum approaching saturation rather than as a limit of either optimizer
(\Cref{app:cross}).

\subsection{RQ2: Which trajectory attributes produce useful model updates?}
\label{sec:rq:recipe}
\label{sec:rq:scale}

\begin{figure}[!htb]
\centering
\includegraphics[width=\linewidth]{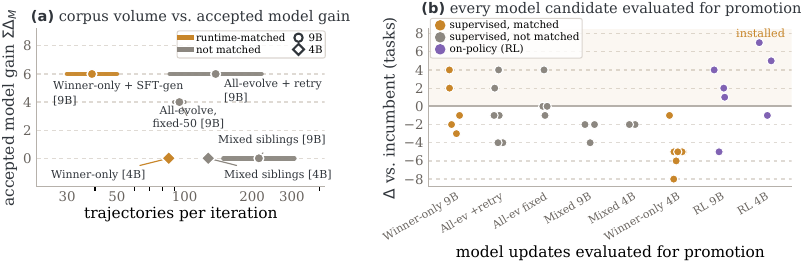}
\caption{\textbf{Trajectory constructions and model-side gain.} \textbf{(a)}
Trajectories per iteration against the accepted model gain (\Cref{tab:recipe}).
\textbf{(b)} Every model update assessed for promotion, shown as its change against the
incumbent, by mixture (\Cref{tab:ledger}).}
\label{fig:rq2}
\end{figure}

\textbf{Trajectory volume alone does not explain model-side gain.} Under
identical tournament search, meta-agent, trainer and promotion protocol,
mixed siblings feeds 149--308 trajectories per iteration and yields
$\Sigma\Delta_M{=}0$, with every update below the incumbent, while winner-only $+$
SFT-gen feeds 30--50 and yields $+6$ through two installed updates
(\Cref{fig:rq2}); substantially more search trajectories therefore do not
guarantee an accepted model update. The comparison holds tournament search,
meta-agent, trainer, and promotion protocol constant while varying the complete
corpus recipe: harness matching, freshness, coverage, prompt conditioning,
per-task cap, and replay ratio (\Cref{sec:res:recipe}). Within this controlled
contrast, runtime matching directly explains why the pooled corpus is dominated
by rollouts from harnesses the promotion procedure later rejects. Moreover, a rollout that is on-policy
for the weights can become off-distribution once the harness changes its
prompts, tools, or observation processing
\citep{tajwar2024}. Consistent with this reading, the all-evolve baselines,
which pool every evolve-set success from a sequential search that proposes one
candidate harness per round and therefore carry no rejected siblings, did
produce accepted updates (\Cref{tab:seqchains}).

\textbf{On-policy data changes what is possible when clean supervised
successes are scarce.} At 4B no supervised construction has produced an
installed model update in \fourBSFTChains{} chains: in the \methodname{}-SFT
chain of \Cref{tab:main} the winner-only filter kept only 2, 3 and 2 matched
trajectories per iteration, below the top-up floor, the corpus size under which
the recipe generates fresh rollouts (\Cref{app:method}), and all three updates were
rejected while a \fourBSFTDH{} harness was installed. The on-policy recipe
needs only reward variance within a group, and moves the same policy
\fourBRLAnchor{}~$\rightarrow$~\fourBRLIthree{} in three stages with no
accepted harness candidate (\Cref{fig:rq2}b). In our 4B runs, offline imitation
therefore yields no accepted update when clean successful trajectories are
this sparse, whereas online reinforcement still obtains a relative reward
signal from the same policy (\Cref{app:chains}).

\subsection{RQ3: How does the harness shape cross-domain transfer?}
\label{sec:rq:general}

\begin{figure}[!htb]
\centering
\includegraphics[width=\linewidth]{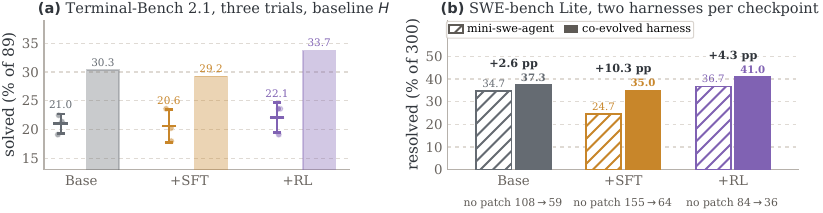}
\caption{\textbf{Transfer under two runtimes.} \textbf{(a)} TB2.1, three
trials under the baseline harness: pass@1 (per-trial solve rate, mean\,$\pm$\,sd
over the three trials; dots) and pass@3 (tasks solved in at least one trial;
bars). \textbf{(b)} SWE-bench Lite resolved rate per checkpoint under the
third-party mini-swe-agent scaffold (hatched) and its co-evolved harness
(solid), with the change in no-patch instances, on which the agent ends
without emitting a patch.}
\label{fig:rq3}
\end{figure}

\textbf{Checkpoint-only gains largely disappear under a foreign runtime.} On
TB2.1 under the baseline harness, three independent trials place the supervised
checkpoint near the base model (\tbSFTThree{} versus \tbBaseThree{} tasks solved
of \NtbTwoOne{}, mean $\pm$ sd over trials) and the reinforcement checkpoint
about one task above the base model on average; at pass@3, the number of tasks
solved in at least one of the three trials, the gap is three tasks
(\Cref{fig:rq3}a). Both differences lie within the trial-to-trial spread despite
10- and 12-task gains on Tmax. This joint domain-and-runtime shift shows that a
substantial fraction of the in-loop gain is conditional on the model--harness
pair.

\textbf{Changing only the runtime materially changes the apparent transfer of
the same checkpoint.} Under mini-swe-agent, a minimal third-party scaffold in
the SWE-agent family \citep{sweagent2024} held fixed across checkpoints, the
three checkpoints resolve \sweBaseResPct\%, \sweSFTResPct\%, and
\sweRLResPct\% of SWE-bench Lite; pairing each with its adopted harness improves
every result, most strongly for SFT (\sweSFTResPct\% to \sweOurSFTResPct\%),
while reducing its no-patch outcomes, instances on which the agent ends without
emitting a patch, from \sweSFTNoPatch{} to \sweOurSFTEmpty{} (\Cref{fig:rq3}b). Because the
weights are fixed within each comparison, this difference isolates the runtime's
contribution. The SFT pair remains below the base pair's \sweOurBaseResPct\%,
whereas reinforcement leads under both harnesses and reaches
\sweOurRLResPct\%; \Cref{app:swe} reports the paired-instance analysis.

\section{Related Work}
\label{sec:related}

\textbf{Harness evolution and model--harness co-evolution.}
Agent-system optimization has progressed from prompt and pipeline tuning \citep{ape2022,opro2023,dspy2023,textgrad2024} to search over agent programs and executable scaffolds \citep{adas2024,dgm2025,alphaevolve2025}. Terminal-agent work optimizes tools, observations, prompts, and recovery under frozen policies \citep{metaharness2026,lemonharness2026,ahe2026,selfharness2026,harnessx2026}, establishing harnesses as a source of capability. Recent systems jointly update both components: Co-Harness alternates harness edits with SFT on updated-runtime trajectories \citep{coharness2026}; Harness-Aware Self-Evolving unifies execution and runtime modification under one RL objective \citep{hase2026}; and EvoTrainer co-evolves policies with training scaffolds, showing that successful trajectories need not be optimal training targets \citep{evotrainer2026}. Post-training dynamics also depend on harness configurations \citep{interplay2026}. These approaches leave routing implicit; \methodname{} makes trajectory filtering, matching, and curriculum scheduling explicit.

\textbf{Data recipes for agent post-training.}
Post-training data selection has progressed from verified self-training and rejection sampling \citep{star2022,rest2023,rft2023,restem2023} to compact curated datasets \citep{lima2023,albalak2024}, on-policy sampling \citep{gkd2023,tajwar2024}, and adaptive task curricula \citep{plr2020}, with recent efforts targeting executable tool use and verifier-grounded terminal feedback \citep{agenttuning2023,agentflan2024,swegym2024,swesmith2025,tmax2026,swerl2025,webrl2024,dapo2025}. Most strategies assume a static runtime, so rollout execution conditions and provenance remain fixed. Model--harness co-evolution instead changes the runtime throughout training, making compatibility and provenance part of data selection. \Cref{app:related} provides an extended discussion.

\section{Conclusion}
\label{sec:conclusion}

We introduced \methodname{}, a harness-aware data recipe for co-evolving
terminal-agent policies and execution harnesses. It routes failures to harness
search, trains policies on verified runtime-matched experience, refreshes the
curriculum, and promotes each component independently. Across supervised and
reinforcement chains, compact matched corpora yielded model gains where much
larger sibling-pooled corpora did not; online reinforcement remained effective
when successes were scarce. Cross-domain evaluations also show that runtime
choice materially shapes checkpoint performance. These results identify
trajectory provenance and model--harness compatibility as central design
variables for agent post-training. \Cref{app:limitations} discusses the
limitations of this study.

\bibliography{refs}
\bibliographystyle{iclr2027_conference}

\newpage
\appendix
\crefalias{section}{appendix}\crefalias{subsection}{appendix}
\etocdepthtag.toc{appendix}
\etocsettagdepth{maintext}{none}
\etocsettagdepth{appendix}{subsection}
\etocsettocstyle{\section*{Appendix}}{}
\makeatletter\patchcmd{\l@section}{1.0em \@plus\p@}{0.45em \@plus\p@}{}{}\makeatother
{\hypersetup{linkcolor=black}\small\tableofcontents}
\clearpage
\section{Extended Related Work}
\label{app:related}

This appendix expands the two paragraphs of \Cref{sec:related}, names the
individual systems the main text groups together, and states, for each
thread, the assumption our recipe changes.

\subsection{Harness and scaffold optimization with frozen weights}
The agent--computer interface is a
design surface in its own right, from custom file viewers and search commands
\citep{sweagent2024} to executable-code action spaces \citep{codeact2024} and
general-purpose platforms \citep{openhands2024,terminalbench2026}. Optimizing that
surface automatically began with prompts \citep{ape2022,opro2023}, grew into
program-level optimization of multi-stage pipelines \citep{dspy2023,textgrad2024}
and into the automated design of whole agents by search over code
\citep{adas2024,dgm2025}, and borrows evolutionary program search over an
archive \citep{funsearch2024,alphaevolve2025} with quality--diversity selection
\citep{noveltysearch2011,mapelites2015,gea2026}. A recent line applies this to
terminal agents specifically, through environment bootstrapping before the
first model call \citep{metaharness2026}, exposure of elapsed and remaining
budget \citep{lemonharness2026}, graph-structured memory \citep{opensage2026},
and observability-driven editing of prompts, tools and middleware
\citep{ahe2026}; \citet{selfharness2026} argue that harnesses are inherently
model-specific and mine verifier-grounded weakness patterns into minimal,
regression-validated edits, and \citet{smallmodels2026} show that frozen-weight
harness optimization recovers a large fraction of frontier performance for
small models. \citet{harnessx2026} supply the typed, hashable substrate we build
on. All of this holds the policy fixed, so its numbers are conditional on a
fixed policy, and it does not examine what the search's trajectories are worth
to a trainer. We reuse the proposal machinery and add a promotion-tested model update that
changes the failure distribution the next search must explain.

\subsection{Model--harness co-evolution and self-improving agents}
Agents that improve
without touching weights accumulate verbal reflections or skill libraries at
test time \citep{reflexion2023,voyager2023}; agents that rewrite their own code
do so under an evaluator they cannot edit \citep{dgm2025}. Several systems now
alternate harness and weight updates. \citet{coharness2026} pair a critic
proposing harness updates with fine-tuning on the improved trajectories;
\citet{hase2026} place task solving and harness editing in one action space
under a shared reinforcement objective with an immutable evaluator, and take
counterfactual edit pairs rather than episodes as the unit of evidence;
\citet{evotrainer2026} co-evolve the policy with the training-side harness,
treat a version transition as the unit of evidence, and note that an
outcome-successful trajectory can be a poor target when it contains looping,
leakage or a verifier mismatch; \citet{interplay2026} show that harness-aware
post-training improves robustness under tool-environment shift. Each reports an
aggregate gain from alternation and, with the partial exception of EvoTrainer's
observation, treats the trajectories flowing between the two optimizers as a
single buffer: whatever the latest harness emitted is what the trainer
consumes. Our focus is the buffer itself: which trajectories should train the
model, under which harness, at which point in the curriculum, and subject to
which promotion rule. We hold the search, trainer, and promotion rule fixed and vary that
construction; under these alternatives, the model channel contributes in one
case but not the other
(\Cref{sec:res:recipe,sec:rq:recipe}). Two related threads sharpen this focus.
Per-trajectory attribution within a training pass \citep{inrunshapley2024}
offers a finer-grained view of rollout value, while measured variance on small
benchmarks \citep{madaan2024} motivates our frozen-split promotion criterion
for every update.

\subsection{Data recipes for agent post-training}
The dominant recipe grows out
of rejection-sampling fine-tuning, in which a model's own verified successes
become its next supervised corpus \citep{star2022,rft2023,rest2023,restem2023}.
It is scaled to agents by harvesting tool-use trajectories from executable
environments, first through instruction-tuning corpora
\citep{agenttuning2023,agentflan2024}, then through verifier-scored trajectory
collection for software and terminal tasks
\citep{swegym2024,swesmith2025,tmax2026,swerl2025} and self-evolving task
curricula for web agents \citep{webrl2024}, optionally followed by
reinforcement learning against verifiable rewards
\citep{ppo2017,grpo2024,tulu2024,dapo2025}. Within this recipe the levers are
well studied. Outcome filtering and per-task caps guard against mode collapse;
on-policy samples are consistently the better target for both preference
tuning and distillation \citep{tajwar2024,gkd2023}; a small, carefully
selected corpus can match a far larger one \citep{lima2023,albalak2024}; and
frontier-driven curricula from reinforcement learning keep the training
distribution at the edge of what the policy can do \citep{plr2020}. All of
this work holds the harness fixed for the duration of training, so a
trajectory's provenance, the prompt and processors it was generated under, is
not a variable the recipe can condition on. Model--harness co-evolution
changes that: the harness moves during training, so the same trajectory can
be on-distribution for one iteration and off-distribution for the next. Our
recipe differs in exactly that respect. It keys the corpus to the fingerprint
of the harness that will run the policy, keeps it current-first with a capped
history so that the curriculum's motion reaches the trainer, retires a task
only once it is solved \emph{and} harvested, and treats a corpus built from
the same bank under a different provenance as a controlled alternative rather
than as more data (\Cref{sec:res:recipe}). In our comparison, the larger
corpus pooled across sibling harnesses yields no accepted model gain, whereas
the smaller harness-matched recipe does (\Cref{sec:rq:recipe}). The on-policy
finding carries over in a specific form: a rollout that is on-policy for the
weights can become off-distribution once the harness changes, because it is
on-distribution only for the model--harness pair that generated it.

\section{Limitations}
\label{app:limitations}

Our evidence comes from a small number of co-evolution chains, each run once.
The promotion split has \NholdoutTmax{} tasks and single evaluations vary by
roughly $\pm2$ tasks (\Cref{app:noise}), so differences of one or two tasks
between chains, including the gap between \methodname{}-SFT and the
mixed-siblings baseline in \Cref{tab:main}, should not be read as rankings;
our claims rest on paired, within-chain contrasts and on the direction of the
accepted gains. The study covers one model family at two scales, one task
source for the loop, and two external benchmarks, so the transfer findings
describe these runtimes rather than runtimes in general. The supervised and
reinforcement recipes were not matched in compute, and harness search depends
on a proprietary meta-agent whose proposals we release only as the accepted
configurations. Finally, the chains stop as the curriculum saturates
(\Cref{sec:rq:relay}); whether a larger task reservoir would extend the gains
is left open.

\section{Experimental Setup}
\label{app:protocol}

\begin{figure}[H]
  \centering
  \includegraphics[width=\linewidth]{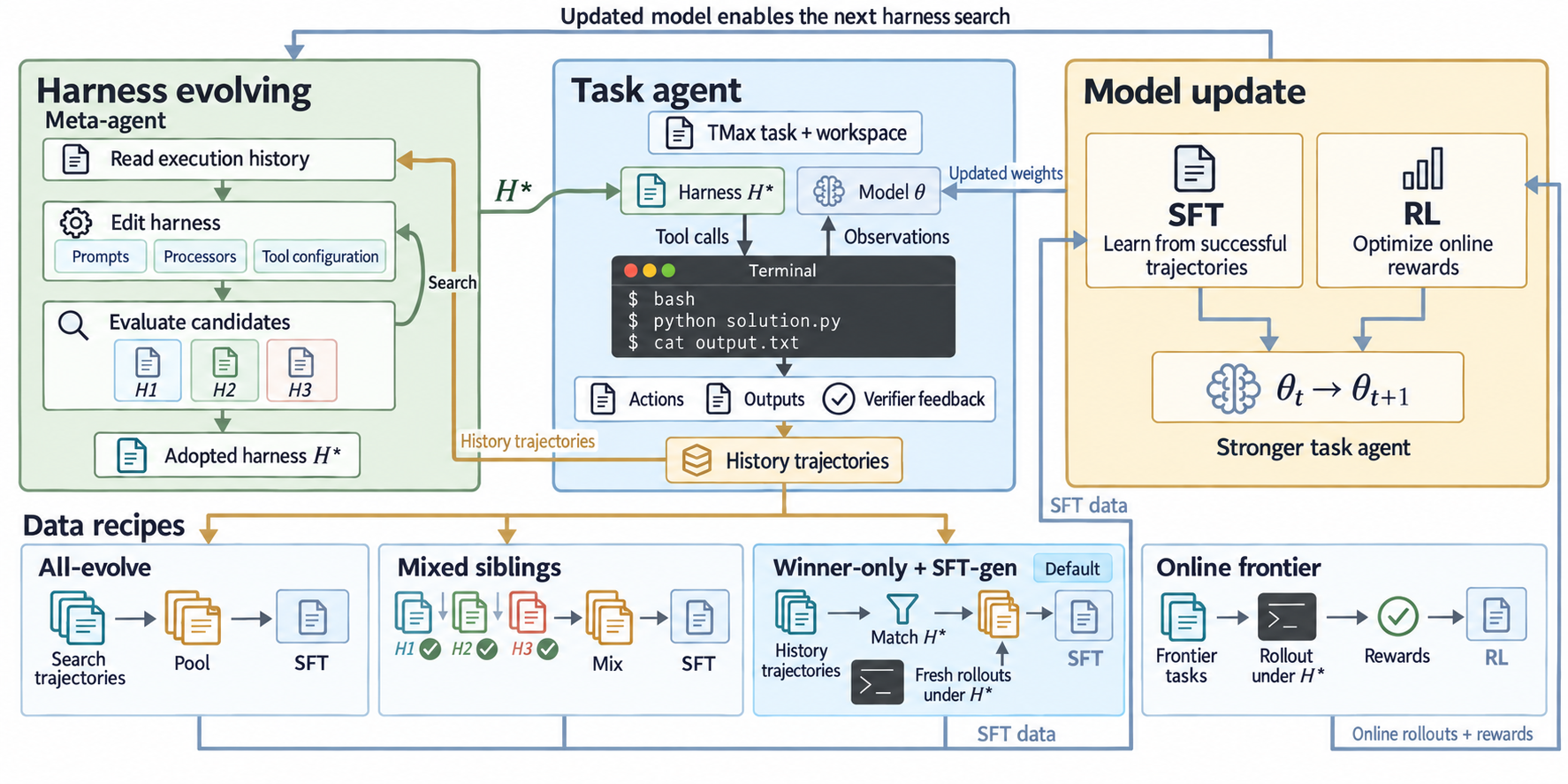}
  \caption{\textbf{Expanded view of the CoTrace workflow.} A meta-agent searches
  for an improved harness $H^*$; the task agent executes it to produce verified
  trajectories; these data support SFT and RL; and the updated model initiates
  the next harness-search round.}
  \label{fig:app:cotrace-workflow}
\end{figure}

\subsection{Chain protocol and reporting rules}
A \emph{chain} is one run of the outer loop from a fixed anchor through $N$
outer iterations. Each iteration executes stages A0~(refresh the curriculum),
A~(search the harness on $\mathcal{E}_t$), B~(evaluate the candidate harness on
$\mathcal{V}$ with the incumbent model), B2~(top up the SFT-generation plane
under the adopted harness), C~(route and build the corpus), D~(train),
optionally D2~(reinforcement stage), and E~(evaluate the candidate model on
$\mathcal{V}$ under the adopted harness). Four rules govern what a chain
reports. (i)~Every score is a pass count on the frozen \NholdoutTmax-task Tmax
promotion split. (ii)~A chain is reported by its anchor and its final
incumbent (the peak for the reinforcement line, with its last measurement
alongside), never by a best intermediate stage. (iii)~Within-chain stage deltas
are more trustworthy than cross-chain absolute scores, because chain anchors
span \anchorNineB{} at 9B and \anchorFourB{} at 4B under an identical protocol. (iv)~A tie is accepted
only when the candidate's own evaluation run was clean, with no missing
results and no tasks in an infrastructure-error status.

\subsection{Task pool and splits}
\label{app:tasks}
\begin{table}[h]
\centering
\caption{Data roles on Tmax \citep{tmax2026}. No task appears in more than one role, and the promotion
split is never optimized against by the harness search, the corpus builder, or
the model update.}
\label{tab:splits}
\small
\begin{tabular}{@{}llll@{}}
\toprule
Role & Tasks & Used by & Optimized against? \\
\midrule
Evolve (rotating) & \NevolveTmax{} / iter & harness search, corpus harvest & yes \\
RL train & $\le$ \NrlTmax{} & online RL rollouts & yes \\
Promotion & \NholdoutTmax{} & both promotion decisions, all reported numbers & \textbf{no} \\
Terminal-Bench 2.1 & \NtbTwoOne{} & transfer reporting only & \textbf{no} \\
\bottomrule
\end{tabular}
\end{table}

Tasks come from Tmax \citep{tmax2026}, an executable terminal-task taxonomy
of roughly \NpoolTmax{} entries. A row is usable only if it carries a non-empty
natural-language description, a final-state test, and a container definition
that converts to a buildable image; rows failing any of these are dropped
before sampling. Each task is a container image, an initial filesystem state,
an instruction, and a programmatic verifier returning a binary reward. The
loop keeps three disjoint planes, as in \Cref{sec:formulation}: a rotating
evolve set $\mathcal{E}_t$ of \NevolveTmax{} tasks, a frozen promotion split
$\mathcal{V}$ of \NholdoutTmax{} tasks, and the remaining taxonomy
$\mathcal{T}$ as the refill reservoir. A leakage check aborts the run if any
evolve, corpus, or RL task id appears in $\mathcal{V}$. Two independent
initial evolve sets are used across chains (a seed-42 draw and a disjoint
alternative draw) so that results are not an artifact of one sample; chains
drawing the alternative set also exclude every id used by earlier chains.

\subsection{Curriculum rotation and execution limits}
From iteration \rotateFrom{}, tasks satisfying \Cref{eq:refresh} are retired and
the set is refilled to \evolveSetSize{} by a domain round-robin over the
taxonomy, excluding $\mathcal{V}$, everything already mastered, and the tasks
being kept. The run aborts rather than evolve against fewer than
\rotateMinSet{} tasks. One measured rotation illustrates the effect: an
iteration kept \rotKept{} unsolved tasks, drew \rotNew{} new ones, retired
\rotRetired{}, and the refreshed set's baseline pass rate fell to
\rotNewR{}; the curriculum is deliberately harder than the one it replaces.
Task-agent rollouts run under a step cap and a per-call token cap
(\Cref{tab:hparams}); evolve-set and promotion-set evaluations use different
container concurrency because the promotion split is larger and more
image-diverse. Container images are built per task; an image-store preflight
sized to the run's own task count aborts the job rather than let a full disk
present itself as a wave of task failures.

\subsection{Compute and cost per lever}
All promotion evaluations run on one 8$\times$H200 node; earlier chains used
8$\times$A100-40GB, on which the reinforcement stage does not fit alongside a
vLLM replica and is disabled. A single \NholdoutTmax-task Tmax promotion
evaluation takes roughly 1.5--2\,h. The \emph{Cost} column of \Cref{tab:main}
is the chain's recorded per-stage wall-clock less its
one-off anchor evaluation, divided by completed outer iterations and
multiplied by the eight GPUs of the node. The composition of that total over
the chains in \Cref{tab:main} is: harness search
\CONFIRMED{14}--\CONFIRMED{57}\%, promotion and stage evaluations
\costEvalShare\%, the reinforcement stage \CONFIRMED{63}\%\ where it runs,
SFT-gen top-up \CONFIRMED{2}--\CONFIRMED{35}\%\ where it runs, and the
supervised gradient step \costGradShare\%. Two things follow. Screening
cheaply before scoring expensively makes the search affordable; moreover,
corpus construction adds little cost relative to the surrounding loop, making
data routing an economical lever for improving the model channel.
\Cref{tab:cost} puts the 9B chains on one budget, in GPU-hours per installed
task. Harness search is cheap and front-loaded: the harness-only chain spends
\totHonly{} GPU-hours in all, \gphHonly{} per installed task, and the
mixed-siblings chain is the next cheapest (\gphMixed{}) because all of its
gain is harness-side. The model channel's cost is evaluation and top-up rather
than the gradient, so the recipe decides whether that spend returns anything;
the retry-enabled chain spent the most per task (\gphRotating{}) because each
rejection bought more search. Reinforcement buys the highest peak at \gphRL{}
GPU-hours per task and the only regression (\gphRLfinal{} once it is counted).
Together with the traces in \Cref{fig:rq1}, this cost profile motivates a
sequence in which the harness evolves first, a matched model update follows,
and reinforcement proceeds only after promotion evaluation.

\begin{table}[h]
\centering
\caption{\textbf{What each lever cost, and what it returned.} The 9B chains of
\Cref{tab:main,tab:seqchains}: completed outer iterations, GPU-hours per iteration, and
GPU-hours per installed task (iterations $\times$ GPU-hours per iteration,
divided by $\Delta$; bracketed: the reinforcement line's last measurement).
Iteration counts and totals are from each chain's per-stage wall-clock log; the
harness-only chain is one iteration of four evolve rounds. A
chain's cost is dominated by evaluation (\costEvalShare\%) and harness search
(\CONFIRMED{14}--\CONFIRMED{57}\%); the gradient step is \costGradShare\%,
SFT-gen top-up \CONFIRMED{2}--\CONFIRMED{35}\%\ and the reinforcement stage
\CONFIRMED{63}\%\ where they run (\Cref{app:protocol}).}
\label{tab:cost}
\footnotesize
\setlength{\tabcolsep}{3.5pt}\renewcommand{\arraystretch}{1.06}
\begin{tabular}{@{}lcccccc@{}}
\toprule
Chain (9B; \Cref{tab:main,tab:seqchains}) & Iters & GPU-h/iter & $\Delta_H$ & $\Delta_M$ & $\Delta$ & GPU-h/task \\
\midrule
Harness only & \itersHonly{}    & \costHonly{}    & $+4$ & ---  & $+4$  & \gphHonly{} \\
All-evolve, fixed-50 & \itersFixed{}    & \costFixed{}    & $+1$ & $+4$ & $+5$  & \gphFixed{} \\
All-evolve $+$ retry & \itersRotating{} & \costRotating{} & $+1$ & $+6$ & $+7$  & \gphRotating{} \\
Mixed siblings & \itersMixed{}    & \costMixed{}    & $+9$ & $0$  & $+9$  & \gphMixed{} \\
\rowcolor{sftrow}
\methodname{}-SFT & \itersWinner{}   & \costWinner{}   & $+4$ & $+6$ & $+10$ & \gphWinner{} \\
\methodname{}-RL & \itersRL{}       & \costRL{}       & $+5$ & $+7$ & $+12$ & \gphRL{} (\gphRLfinal{}) \\
\bottomrule
\end{tabular}
\end{table}

\subsection{Reproducibility and evaluation noise}
\label{app:noise}
Task-agent decoding is greedy ($T=0$); curriculum rotation and corpus sampling
are seeded (\refillSeed{}). Harness search invokes a proprietary meta-agent,
and we release the accepted configuration from every iteration so that each
reported promotion evaluation can be reproduced directly without rerunning
the search.

The \NholdoutTmax-task split is consulted by both promotion decisions; we
therefore call it the \textbf{promotion} split throughout and do not treat it as
a test set. It provides the
common reference for within-chain dynamics, while the external benchmarks of
\Cref{sec:rq:general}, which no stage of the loop consults, measure transfer.

Three views of single-evaluation noise agree. First, chain anchors (the same
base checkpoint under the same stock harness, evaluated at the start of
independent chains) span \anchorNineB{} pass counts at 9B and \anchorFourB{} at
4B (\Cref{tab:ledger}). Second, the same evolved harness on the same frozen 4B
weights scored \reEvalFourBLo{} in the per-round curve of the harness-only
control and \reEvalFourBHi{} in its promotion evaluation. Third, in the
reinforcement line the paired net change and the difference of separately
executed stage evaluations differ by one to two tasks at every recorded stage
(\Cref{tab:flips}). Together they place single-evaluation noise at roughly
$\pm2$ tasks, with a worst observed excursion of five tasks (4B). These
estimates characterize the scale of single-evaluation variation and motivate
the paired, within-chain analysis used for the mechanism claims in
\Cref{sec:analysis}.

\subsection{Hyperparameters}
\label{app:hparams}
\Cref{tab:hparams} lists the search, corpus-construction, training, and
promotion settings used in every reported chain.
\begin{table}[h]
\centering
\small
\caption{Full configuration. The reinforcement-stage settings are those of the
line in \Cref{fig:trajectories}(b); \Cref{tab:rlhparams} gives its rollout
budget.}
\label{tab:hparams}
\begin{tabular}{@{}llll@{}}
\toprule
\multicolumn{2}{@{}l}{\textbf{Harness search}} & \multicolumn{2}{l}{\textbf{SFT}} \\
Tournament              & $2\times\tourN$ candidates & Method & LoRA \\
Extra rounds on failure & \CONFIRMED{2}        & LoRA rank / $\alpha$ & \CONFIRMED{32} / \CONFIRMED{64} \\
Max SFT retries         & \CONFIRMED{2}        & LoRA dropout        & \CONFIRMED{0.05} \\
Meta-agent step cap     & \CONFIRMED{200}      & Targets             & q,k,v,o,gate,up,down \\
Meta-agent wall clock   & \CONFIRMED{3600}\,s  & Epochs              & \CONFIRMED{2} \\
Evolve set size         & \NevolveTmax{}       & Learning rate       & \CONFIRMED{$2\times10^{-5}$} \\
Rotation start          & iteration \CONFIRMED{2} & Schedule / warmup & linear / \CONFIRMED{0.03} \\
Evolve set mix (fail/pass) & \mixFail/\mixPass & Max seq.\ length    & \CONFIRMED{4096} \\
Screen abort threshold  & \screenErrWarn\%\ errored & Promotion rule & $w > \ell$; ties if clean \\
\multicolumn{2}{@{}l}{\textbf{Population search (screened variant)}} & \multicolumn{2}{l}{}     \\
Fan-out width $N$       & \fanoutN{}           &                     &        \\
Survivors $k$ / concurrency & \fanoutKeep{} / \fanoutConc{} &        &        \\
Explore period $E$      & \exploreEvery{}      &                     &        \\
Novelty neighbours      & \knnK{}              &                     &        \\
Probe set (solved/failed) & \probeSolved/\probeUnsolved &            &        \\
Probe regressions tolerated & \probeMaxReg{}   &                     &        \\
\multicolumn{2}{@{}l}{\textbf{Corpus construction}}    & Effective batch     & \CONFIRMED{8} \\
Max trajectories        & \CONFIRMED{100}      & Precision           & BF16 \\
Max per task            & \CONFIRMED{3} (default recipe: 1) & Loss   & completion-only \\
Tool-call range         & \CONFIRMED{[2, 60]}  &                     & \\
Min trajectories        & \CONFIRMED{20}       & \multicolumn{2}{l}{\textbf{Online RL}} \\
Train pairs / iteration & \CONFIRMED{526--948} (default recipe) & Objective & DPPO, binary TV $\delta{=}\rlTV{}$ \\
SFT-gen target tasks    & \sftGenTarget{}      & Episodes            & \rlEpisodes{} \\
History cap             & \histCap\%           & Train tasks         & \rlTasks{} \\
\multicolumn{2}{@{}l}{\textbf{Evaluation}}     & Advantages          & centered, group-relative \\
Decoding                & greedy               & Group shape / LR    & \rlGroup{} / \rlLR{} \\
Per-call token cap      & \CONFIRMED{4096}     & Reference KL $\beta$ & \rlBeta{} \\
Promotion tolerance     & \CONFIRMED{0} (ties allowed if clean) & LM head / response & FP32 / \rlRespLen{} \\
Held-out concurrency    & \CONFIRMED{2}        &                     & \\
\bottomrule
\end{tabular}
\end{table}

\section{Method Details}
\label{app:method}

\subsection{Co-evolution loop in pseudocode}
\label{app:algorithm}

The complete workflow is organized into four modules. \Cref{alg:cotrace}
evolves and promotes the harness; \Cref{alg:data-recipe} routes traces and
refreshes the curriculum; and \Cref{alg:sft-update,alg:rl-update} give the two
model-update alternatives. Together they implement one outer iteration of
\Cref{alg:cotrace-simple}. In all four modules, the promotion split
$\mathcal{V}$ remains disjoint from search and training data. We write
$\textsc{Promote}_{\mathcal{V}}(c,i\mid q)$ for a paired comparison that holds
component $q$ fixed and returns candidate $c$ when it wins more tasks than it
loses (or ties with a clean run), and incumbent $i$ otherwise.

\begin{algorithm}[tbp]
\caption{Harness evolution and promotion}
\label{alg:cotrace}
\label{alg:harness-evolve}
\small
\begin{algorithmic}[1]
\Require policy $\theta_t$, harness $H_t$, evolve set $\mathcal{E}_t$,
failure clusters $\mathcal{D}^{H}_t$, promotion split $\mathcal{V}$
\State Initialize archive $\mathcal{A} \gets \{H_t\}$ and parent $H^{\mathrm{par}} \gets H_t$
\For{two tournament generations}
  \State assign recurring failure clusters as distinct proposal foci
  \State $\mathcal{P} \gets \textsc{MetaAgent}(H^{\mathrm{par}}, \mathcal{D}^{H}_t)$
  \State discard candidates failing structural, impact, probe, or admissibility checks
  \State evaluate each survivor on $\mathcal{E}_t$ with $\theta_t$ fixed; append scores and traces to $\mathcal{A}$
  \State $H^{\mathrm{par}} \gets$ highest-mean candidate eligible under tolerance $\innerTol$
\EndFor
\State $H^{\mathrm{cand}} \gets$ highest-mean candidate in $\mathcal{A}$
\State $H_{t+1} \gets \textsc{Promote}_{\mathcal{V}}(H^{\mathrm{cand}},H_t\mid\theta_t)$
\State \Return adopted harness $H_{t+1}$ and all search traces
\end{algorithmic}
\end{algorithm}

\begin{algorithm}[tbp]
\caption{Route traces and refresh the data recipe}
\label{alg:data-recipe}
\small
\begin{algorithmic}[1]
\Require bank $\mathcal{B}_{0:t}$, policy $\theta_t$, adopted harness $H_{t+1}$,
evolve set $\mathcal{E}_t$, task pool $\mathcal{T}$, promotion split $\mathcal{V}$
\State $\mathcal{D}^{H}_t \gets$ recurring earliest-failure clusters; exclude environment and grader faults
\State $\mathcal{S}_t \gets$ verified, uninterrupted successes whose harness fingerprint matches $H_{t+1}$
\State inject the system prompt of $H_{t+1}$ into every retained training example
\If{$|\mathrm{tasks}(\mathcal{S}_t)| < \sftGenTarget$}
  \State roll out $(\theta_t,H_{t+1})$ on fresh tasks and add verified successes
\EndIf
\State quality-rank and deduplicate; keep current data first, at most \perTaskDefault{} trace per task,
$|\mathcal{S}_t|\le\corpusMax$, and history $\le\histCap\%$
\State $\mathcal{M}_t \gets \{x\in\mathcal{E}_t: x\text{ is solved and represented in }\mathcal{S}_t\}$
\State $\mathcal{E}_{t+1} \gets \textsc{RefillByDomain}(\mathcal{E}_t\setminus\mathcal{M}_t,
\mathcal{T}\setminus\mathcal{V})$ to \evolveSetSize{} tasks
\State \Return failure evidence $\mathcal{D}^{H}_t$, SFT corpus $\mathcal{S}_t$, and $\mathcal{E}_{t+1}$
\end{algorithmic}
\end{algorithm}

\begin{algorithm}[tbp]
\caption{Supervised policy update}
\label{alg:sft-update}
\small
\begin{algorithmic}[1]
\Require policy $\theta_t$, adopted harness $H_{t+1}$, corpus $\mathcal{S}_t$,
promotion split $\mathcal{V}$
\If{$\mathcal{S}_t=\varnothing$}
  \State \Return $\theta_t$
\EndIf
\State $\theta' \gets \textsc{LoRA-SFT}(\theta_t,\mathcal{S}_t)$
  \Comment{completion-only loss; continue the incumbent adapter}
\State $\theta_{t+1} \gets \textsc{Promote}_{\mathcal{V}}(\theta',\theta_t\mid H_{t+1})$
\If{the candidate is not promoted and retries remain}
  \State collect fresh search traces, rebuild $\mathcal{S}_t$, and retry with new data
\EndIf
\State \Return $\theta_{t+1}$
\end{algorithmic}
\end{algorithm}

\begin{algorithm}[tbp]
\caption{Online reinforcement update}
\label{alg:rl-update}
\small
\begin{algorithmic}[1]
\Require policy $\theta_t$, adopted harness $H_{t+1}$, task pool $\mathcal{T}$,
promotion split $\mathcal{V}$
\State sample \rlTasks{} training tasks outside $\mathcal{V}$:
\rlFrontierPct\%\ from the unsolved frontier, the rest domain-balanced
\State collect grouped online rollouts under $(\theta_t,H_{t+1})$ and score them with binary verifiers
\State discard zero-variance groups and refill by active sampling
\State $\theta' \gets \textsc{DPPO}(\theta_t,\text{rollouts})$
  \Comment{binary-TV trust region; outcome-only rewards}
\State $\theta_{t+1} \gets \textsc{Promote}_{\mathcal{V}}(\theta',\theta_t\mid H_{t+1})$
\State \Return $\theta_{t+1}$
\end{algorithmic}
\end{algorithm}

\subsection{Curriculum refresh and baseline recipes}
\label{app:refresh}
At the end of iteration $t$ the mastered set and the next evolve set are
\begin{align}
    \mathcal{M}_t
    &= \left\{ x \in \mathcal{E}_t : \mathrm{solved}_t(x) \wedge \mathrm{harvested}_t(x) \right\},
    \label{eq:mastered}
    \\
    \mathcal{E}_{t+1}
    &= \bigl(\mathcal{E}_t \setminus \mathcal{M}_t\bigr) \cup \mathrm{Refill}\!\left(\mathcal{T}_{\mathrm{pool}} \setminus (\mathcal{V} \cup \mathcal{M}_{0:t} \cup \mathcal{E}_t)\right),
    \label{eq:refresh}
\end{align}
where $\mathrm{solved}_t(x)$ denotes success by the incumbent and
$\mathrm{harvested}_t(x)$ inclusion in $\mathcal{D}^M_{0:t}$; the conjunction
retires a task only after it has provided verified training signal. The
evolve set is replenished to \evolveSetSize{} tasks by domain-stratified
sampling from $\mathcal{T}_{\mathrm{pool}}$, falling back to empirical success
during online reinforcement learning.

The two baseline recipes of \Cref{tab:main,tab:recipe} are defined as
follows. \emph{All-evolve} pools verified search successes up to $c_r$ per
task without conditioning on fingerprints or synthesizing extra rollouts; it
was run under sequential search on a static evolve set (\emph{Seq.,
fixed-50}) and on a rotating set with retry upon rejection (\emph{Seq.,
rotating-50}, \emph{all-evolve $+$ retry}), and is reported in
\Cref{tab:seqchains}. \emph{Mixed siblings} aggregates verified successes
from every candidate harness scored during the tournament at an expanded
per-task cap, keeping the original exploratory prompts, so its volume does
not depend on which candidate the promotion procedure adopted.

\subsection{Baseline harness configuration}
\label{app:baseline}

Every harness gain in the paper is measured against the hand-written scaffold
below, which is already the product of several rounds of manual tuning on
Terminal-Bench. It is a HarnessX configuration \citep{harnessx2026}: a
\texttt{tool\_registry} (the shell tool is the sole action) and an ordered
list of processors, each subscribed to a hook of the agent loop. Reporting
harness-search gains against a thin scaffold instead of this baseline would
roughly double the apparent effect. The meta-agent may edit the processor
list, the tool registry and the accompanying system-prompt file
(\Cref{app:prompts}); the evaluator, sandbox provider and tracer are injected
by the runner and lie outside every editable surface.

\begin{configbox}[nobreak=false,frametitle={Baseline harness configuration (comments and hook bindings elided; every processor binds to every hook)}]
\begin{lstlisting}[style=plainbox]
tool_registry:
  builtin:
  - Bash
  custom: []
processors:
- _target_: context.system_prompt.SystemPromptProcessor
  system_builder:
    _target_: tmax.prompt_builder.SiblingSystemPromptBuilder
- _target_: context.env_context_injector.EnvironmentContextInjector
  working_dir: /home/user
  constraints: {}
  header: Environment
  max_tree_lines: 20
  non_interactive: true
  inject_integrity_rules: true
  show_project_dir: false
- _target_: control.tool_call_correction.ToolCallCorrectionLayer
  tool_schemas: {}
- _target_: tb2.TaskTimeReminderProcessor
  warn_at:
  - 0.7
  - 0.9
- _target_: tmax.processors.length_recovery.LengthTruncationRecoveryProcessor
  repeat_threshold: 2
  head_chars: 1200
  tail_chars: 600
- _target_: control.compaction.CompactionProcessor
  token_threshold: 140000
  message_threshold: 100
  retention_window: 6
  eviction_fraction: 0.5
  summarize_key: summarize
  preserve_first_message: true
- _target_: control.parse_retry.ParseRetryProcessor
  max_consecutive_errors: 1
- _target_: tb2.PostCompactionRefreshProcessor
  drop_threshold: 5
- _target_: control.bg_install_guard.BgInstallGuard
- _target_: tb2.CustomEditToolProcessor
  threshold: 7
- _target_: tb2.CustomSelfVerifyProcessor
\end{lstlisting}
\end{configbox}

Reading the list top to bottom: the system-prompt processor reads the prompt
file next to the configuration; the environment-context injector reports the
working directory, a 20-line file tree, available interpreters and package
managers, and the integrity rules; the tool-call correction layer repairs
malformed calls; the task-time reminder fires at 70\% and 90\% of the step
budget; the length-truncation recovery processor breaks \texttt{max\_tokens}
repetition loops after two repeats; the compaction processor summarizes at
140k tokens or 100 messages with a retention window of six; the parse-retry
processor gives one retry after a malformed response; the post-compaction
refresh re-injects context dropped by compaction; the background-install guard
prevents detached package installs from being mistaken for progress; and the
custom edit tool and self-verification processor add a file-edit affordance
and a final check before the agent stops.

\subsection{Screens and admissibility checks in harness search}
\label{app:screens}

\paragraph{Screens (heuristic ranking).}
\textbf{Structural} diffs each candidate against the parent and fingerprints
the changeset using the semantic hashes of \Cref{sec:method:route},
dropping empty changesets, intra-batch duplicates, and signatures the archive
already rejected, all without a model call. \textbf{Predicted impact} has the
meta-model rank survivors and commit, per candidate, which tasks the edit
should newly solve and which already-solved tasks it puts at risk; because
those predictions are stored on the node, a later round can score the
proposer's calibration rather than trusting its ranking indefinitely.
\textbf{Mini-eval probe} runs survivors concurrently on the mixed probe set and
drops any candidate breaking more than \probeMaxReg{} parent-solved task.

The cascade fails open: an erroring screen passes its top-$k$ input through
rather than emptying the round, and a candidate whose probe hit an
infrastructure fault stays alive unscored rather than being penalized for the
cluster's behavior, the same principle \routername{} applies to trajectories.
Every drop records which screen fired and why, because a screen that kills
every candidate and one that kills none are both bugs that an aggregate score
hides. Probe tasks are chosen deterministically so scores stay comparable across
rounds of a resumed campaign.

\paragraph{Hard admissibility checks.}
\textbf{(1)~Canonicalize}: the emitted YAML must bind to a live harness object
and round-trip. \textbf{(2)~Dry-fire}: every authored processor and tool is
invoked once with schema-derived dummy input; exceptions fail the round.
\textbf{(3)~Contract}: authored processors are checked against the mutation
contract of the hook they subscribe to, so a machine-authored processor cannot
silently rewrite history and produce a trajectory that is not a valid training
target. \textbf{(4)~Non-repetition}: re-proposing a previously reverted edit
requires an explicit rationale in the journal. \textbf{(5)~Evidence}: each
candidate must name a failure cluster, a causal mechanism, an exact patch, a
predicted-affected task set, and a rollback condition. \textbf{(6)~Replay}: the
config boots through the real run loop on one synthetic task; any crash fails
the round. Checks 1--3 and 6 cost seconds against roughly
\costEvolveHours{}~GPU-hours for a scoring round, which is what makes a wide
fan-out affordable. Every round appends an auditable journal entry (failure
hypothesis, trajectory evidence, proposed change, expected gain, regression
risk, rollback condition), which is also the substrate for the non-repetition
check.

\subsection{Failure attribution and routing}
\label{app:routing}

The router uses deterministic trace features. It first localizes the
\emph{earliest unrecovered failure}: a failed tool observation counts as
recovered if a later observation for the same command family succeeds, and all
later events become downstream symptoms. It then classifies the critical event
and attaches a confidence, quarantining scores below \routerConfThresh{}.
Environment and grader faults have dedicated destinations and never enter the
corpus or search evidence. \Cref{tab:routerrules} gives the rules; the key
contrast is between tool-wrapper or error-not-surfaced failures and repeated
actions, which are indistinguishable by reward but opposite in prescription.

\begin{table}[h]
\centering
\caption{Attribution rules of the router, in application order, with the
confidence each assigns and the destination it selects. Successes are routed
too: a success that recovered from an earlier failure becomes a
recovery-slice demonstration.}
\label{tab:routerrules}
\footnotesize
\setlength{\tabcolsep}{4pt}\renewcommand{\arraystretch}{1.08}
\begin{tabular}{@{}llcl@{}}
\toprule
Observed pattern & Attribution & Conf. & Destination \\
\midrule
No model input tokens consumed (environment never started) & \textsc{env} & 0.97 & environment repair \\
Grader or verifier exception & \textsc{grader} & 0.90 & benchmark repair \\
Session log absent & --- & 0.00 & quarantine \\
Clean success & \textsc{model} & 0.80 & full-trajectory SFT \\
Success after a recovered failure & \textsc{model} & 0.80 & recovery-slice SFT \\
Failed with no error, step budget exhausted & \textsc{ambiguous} & 0.45 & quarantine \\
Failed with no error, stopped early & \textsc{model} & 0.60 & model training \\
Disk, DNS, or package-manager failure in the observation & \textsc{env} & 0.85 & environment repair \\
Tool wrapper rejected or suppressed the call & \textsc{tool} & 0.85 & harness evolution \\
Error never surfaced to the model (empty observation) & \textsc{tool} & 0.70 & harness evolution \\
Visible error, identical command repeated & \textsc{model} & 0.80 & model training \\
Visible error, no effective recovery & \textsc{model} & 0.65 & model training \\
\bottomrule
\end{tabular}
\end{table}

\subsection{Prompts and meta-agent briefs}
\label{app:prompts}

\paragraph{Task-agent system prompt.} The baseline scaffold ships a
deliberately minimal five-line prompt. It is one of the three editable
surfaces, so a chain may replace it; \Cref{app:converge} reports that the two
highest-scoring harnesses kept it unchanged. Every training example carries
the system prompt of the harness whose fingerprint it was generated under, so
a trajectory harvested under an evolved prompt is never trained as though it
came from the stock one (\Cref{sec:res:recipe} measures what happens when that
alignment is dropped).

\begin{promptbox}[frametitle={Task-agent system prompt (stock Tmax prompt)}]
\begin{lstlisting}[style=plainbox]
You are a terminal coding agent solving a single Linux task.
Use the Bash tool to inspect the environment, edit files, and run commands.
Work under /home/user unless the instruction says otherwise.
When the task is complete, stop calling tools and briefly confirm what you did.
Do not ask questions -- act.
\end{lstlisting}
\end{promptbox}

\paragraph{Meta-agent brief.} The meta-agent receives a generated brief rather
than a fixed prompt. Its sections, in order: the assigned per-sibling
\textbf{focus}; \textbf{pivot harnesses} to read before proposing; the
\textbf{global optimization constraint} (net gain, newly solved tasks weighed
against regressions, \probeNet{}); the \textbf{evolve brief} (current config
path, trajectory directory, output directory, journal memo, and a
machine-rendered lever scoreboard with per-task history and recent
changesets); the \textbf{deliverables} it must write (a harness configuration file and
a system-prompt file in the round's output directory); a
\textbf{self-validation} checklist to run before ending its turn; and a
\textbf{decision contract} requiring, for each candidate, a named failure
cluster, a causal mechanism, an exact patch, a predicted-affected task set, and
a rollback condition. The predicted-affected set is what the probe screen later
falsifies, and the journal supplies the evidence for the non-repetition check. The focus is
what makes the $N$ sibling proposals of a round distinct: each is anchored to a
different failing task where enough failures exist, and otherwise to a
different lever, crossed with an edit style once the levers run out.

\begin{briefbox}[frametitle={Meta-agent focus assignments: failure anchors, six levers, three edit styles}]
\begin{lstlisting}[style=plainbox]
Failure-anchored focus (assigned first, one failing task per sibling):
  Task `<task id>` fails. Read that task's trajectory in `trajectories_dir` first and diagnose why before proposing anything. Fix the harness capability the failure exposes, not the task.
  Observed: <one-line failure digest from the router>

Generic levers (fill the remaining siblings, best-yielding first):
  1. Recovery from tool errors. Find where the agent hit a failing tool call and then repeated it or gave up. Change the harness so the failure is surfaced legibly and a different approach is attempted.
  2. Context hygiene. Find where the agent lost track of earlier findings, re-read files it had already read, or drowned in tool-result noise. Change what the harness keeps, summarises, or filters.
  3. System prompt specificity. Find instructions the agent demonstrably failed to follow, or a missing instruction whose absence explains a failure. Edit the prompt template, not the tool set.
  4. Tool surface. Find a task where the available tools forced an awkward or many-step workaround. Add, remove, or re-describe a tool so the direct path exists.
  5. Stopping and verification. Find where the agent declared success without checking, or burned its budget after the work was already done. Change the harness's completion criteria.
  6. Step budget allocation. Find where the agent ran out of steps mid-task or wasted early steps on exploration that did not pay off. Change pacing, reminders, or planning structure.

Edit style (crossed with the lever when the fan-out exceeds six):
  a. prefer REMOVING or simplifying an existing mechanism over adding a new one; if something in the harness is actively getting in the way, cutting it is a valid and often stronger fix.
  b. prefer the SMALLEST edit that could possibly work; a one-line prompt or parameter change that is clearly attributable beats a broad rewrite whose effect cannot be isolated.
  c. prefer a STRUCTURAL change (a new processor or tool) over prompt wording, if the evidence shows the agent knew what to do but had no mechanism to do it.
\end{lstlisting}
\end{briefbox}

Under the tournament configuration every focus is followed by the
generalization contract below, and the last sibling of a round with two or
more live lineages is instead given the synthesis focus, which asks it to
combine mechanisms from the pivot harnesses rather than propose a new one.

\begin{briefbox}[frametitle={Evidence anchor, generalization contract, and synthesis focus (tournament brief)}]
\begin{lstlisting}[style=plainbox]
Use failing task `<task id>` only as the initial evidence anchor. Read its trajectory, then search for the same failure class in other trajectories before editing.

Generalization contract. Treat the named task as evidence, not as the
target. First state the reusable failure class in terms observable by the
agent (tool result, context state, progress, or verification state). Inspect
at least two other trajectories for supporting or counter-evidence when they
exist. The harness edit must trigger from that general state: do not mention
task IDs, benchmark paths, filenames, expected answers, or task-specific
content in the prompt, processor, or tool. Explain why the mechanism should
help unseen tasks and identify which already-solved task class it could hurt.
If the evidence supports only one task, prefer a no-op over a special case.

Complementary synthesis. Read every pivot config and trajectory directory
in the pivot table. Combine only mechanisms whose per-task coverage is
complementary: preserve a mechanism that uniquely solves tasks, and use another
pivot to repair its unique regressions. Resolve conflicting prompts/processors
instead of blindly concatenating them. The result must still satisfy the
generalization contract and contain no task-specific trigger.
\end{lstlisting}
\end{briefbox}

\paragraph{Reinforcement-stage prompt.} Online rollouts run under the adopted
harness's processors, but the reinforcement trainer imposes its own action
format: one \texttt{bash} tool call per turn, preceded by a reasoning section.
The prompt below is the one whose absence produced the zero-reward runs of
\Cref{app:rl}.

\begin{promptbox}[frametitle={Reinforcement-stage system prompt}]
\begin{lstlisting}[style=plainbox]
You are a helpful assistant that can interact with a computer.

Your response must include a THOUGHT section before your action where you
explain your reasoning. After the THOUGHT, you must call the `bash` tool
with EXACTLY ONE bash command (multiple commands chained with `&&` or `||`
count as a single action).

Failure to follow these rules -- calling no tool, calling a tool other than
`bash`, or omitting the THOUGHT -- will cause your response to be rejected.
\end{lstlisting}
\end{promptbox}

\paragraph{SWE-bench Lite agent prompt.} The transfer evaluation of
\Cref{sec:rq:general} runs the checkpoints inside mini-swe-agent with a
text-based action format, so a turn that does not contain exactly one action
receives the format-error message below and, after repeated misses, an empty
patch. This is the mechanism behind the \sweSFTFormat{} format failures of the
supervised checkpoint against \sweRLFormat{} for the reinforcement one.

\begin{promptbox}[frametitle={SWE-bench Lite: mini-swe-agent system prompt and format-error message}]
\begin{lstlisting}[style=plainbox]
[system]
You are a helpful assistant that can interact multiple times with a computer shell to solve programming tasks.
Your response must contain exactly ONE bash command inside an <mswea_bash_command> block.

Include a THOUGHT section before your command where you explain your reasoning process.
Format your response as shown in <format_example>.

<format_example>
THOUGHT: Your reasoning and analysis here

<mswea_bash_command>your_command_here</mswea_bash_command>
</format_example>

Failure to follow these rules will cause your response to be rejected.

[format-error message, sent when a turn does not contain exactly one action]
Format error:

<error>
{{error}}
</error>

Please always provide EXACTLY ONE action in an <mswea_bash_command> block, found {{actions|length}} actions.

<response_example>
THOUGHT: briefly explain what you will do

<mswea_bash_command>ls -la</mswea_bash_command>
</response_example>

step limit 250; temperature 0; max 4,096 tokens per turn; working directory /testbed.
\end{lstlisting}
\end{promptbox}

\section{Additional Experimental Results}
\label{app:chains}

\subsection{Promotion ledger of every chain}
\Cref{tab:ledger} opens up every chain of \Cref{tab:main} one promotion decision at
a time. Reading it row-wise recovers the $\Delta_H$ and $\Delta_M$ columns of
\Cref{tab:main} and the accepted-update counts of \Cref{tab:recipe};
\Cref{fig:rq1}(a) and \Cref{fig:traces}(b) plot two of its chains. Reinforcement
stages follow the same model-promotion rule as supervised ones; in the 9B line every
reinforcement checkpoint up to the third improved the incumbent and was
installed, and the fourth scored below the incumbent and was rejected, which is
the measurement reported as the line's last.

\begin{table}[h]
\centering
\caption{Promotion ledger for every chain of \Cref{tab:main}, in decision order.
$H$: candidate harness scored with the incumbent model (stage B); $M$:
candidate checkpoint scored under the adopted harness (stage E). Bold marks a
change the promotion procedure installed; a slash separates retries after additional evolve
rounds. The ledger is consistent with \Cref{tab:main,tab:recipe}.}
\label{tab:ledger}
\small
\setlength{\tabcolsep}{5pt}
\begin{tabular}{@{}llrl@{}}
\toprule
Chain (\Cref{tab:main,tab:seqchains}) & Iter & Before & Decisions, in order \\
\midrule
Harness only              & 1--4 & 77 & best candidate per round: 80, 78, 80, 81 (no ratchet) \\
\addlinespace[2pt]
All-evolve, fixed-50      & 1 & 77 & $H$ 75 reject;\; $M$ \textbf{81} accept \\
                          & 2 & 81 & $H$ \textbf{82} accept;\; $M$ 82 / 81 / 82 reject \\
\addlinespace[2pt]
All-evolve $+$ retry      & 1 & 77 & $H$ \textbf{78} accept;\; $M$ \textbf{82} accept \\
                          & 2 & 82 & $H$ 81 reject;\; $M$ 81 reject / \textbf{84} accept \\
                          & 3 & 84 & $H$ 83 reject;\; $M$ 80 / 83 / 80 reject \\
\addlinespace[2pt]
Mixed siblings            & 1 & 77 & $H$ \textbf{83} accept;\; $M$ 81 reject \\
                          & 2 & 83 & $H$ \textbf{86} accept;\; $M$ 84 reject \\
                          & 3 & 86 & $H$ 80 reject;\; $M$ 82 reject \\
\addlinespace[2pt]
\methodname{}-SFT         & 1 & 78 & $H$ 76 reject;\; $M$ \textbf{82} accept \\
                          & 2 & 82 & $H$ \textbf{86} accept;\; $M$ 84 reject \\
                          & 3 & 86 & $H$ 82 reject;\; $M$ 85 reject \\
                          & 4 & 86 & $H$ 80 reject;\; $M$ 83 reject \\
                          & 5 & 86 & $H$ 82 reject;\; $M$ \textbf{88} accept \\
\addlinespace[2pt]
\methodname{}-RL          & 1 & 78 & $H$ 78 tie (\flipHone), carried;\; RL \textbf{79} accept \\
                          & 2 & 79 & $H$ \textbf{80} accept;\; RL \textbf{84} accept \\
                          & 3 & 84 & $H$ \textbf{86} accept;\; RL \textbf{88} accept \\
                          & 4 & 88 & $H$ \textbf{90} accept (\flipHfour);\; RL 85 reject \\
\midrule
4B, harness only          & 1 & 65 & R1--R5: \fourBHoRounds{};\; $H$ \textbf{66} (R3) accept \\
4B, mixed siblings        & 1 & 64 & $H$ 64 reject (unclean tie);\; $M$ 62 reject \\
                          & 2 & 64 & $H$ 60 reject;\; $M$ 62 reject \\
4B, \methodname{}-SFT      & 1 & 65 & $H$ not adopted;\; $M$ 60 reject \\
                          & 2 & 65 & $H$ 63 reject;\; $M$ 60 reject \\
                          & 3 & 65 & $H$ \textbf{69} accept;\; $M$ 63 reject \\
4B, winner-only, 1 iter.\ & 1 & 67 & $H$ \textbf{68} accept;\; $M$ 66 reject \\
4B, winner-only, 3 iter.\ & 1--3 & 67 & $M$ 62;\; $H$ 63 / $M$ 59;\; $H$ 64 / $M$ 62, all rejected \\
4B, mixed, 3 iter.\ & 1--3 & 64 & $H$ 66 / $M$ 65;\; $M$ 61;\; $H$ 56 / $M$ 55; none retained \\
4B, \methodname{}-RL       & 1 & 67 & $H$ not adopted;\; RL \textbf{74} accept \\
                          & 2 & 74 & $H$ 74 tie, not adopted;\; RL 73 reject \\
                          & 3 & 74 & $H$ 68 reject;\; RL \textbf{79} accept \\
\bottomrule
\end{tabular}
\end{table}

\subsection{Sequential-search chains}
\label{app:seqchains}
Two 9B chains predate the tournament search and use the all-evolve recipe
under sequential search (\Cref{app:refresh}); they are not controlled
comparisons against the tournament rows of \Cref{tab:main} and are listed in
\Cref{tab:seqchains}. Both gain mostly through the model channel, which is
the boundary condition noted in \Cref{sec:rq:recipe}: pooled search
trajectories without harness matching can produce accepted updates when the
search is sequential and the pool is not dominated by rejected siblings.

\begin{table}[h]
\centering
\caption{The two sequential-search chains: promotion-split scores and gains
as in \Cref{tab:main}, corpus per iteration as in \Cref{tab:recipe}. Pooled
search trajectories, up to 3 per task; retry adds evolve rounds and rebuilds
the corpus after a rejected model update.}
\label{tab:seqchains}
\scriptsize
\setlength{\tabcolsep}{3pt}\renewcommand{\arraystretch}{1.08}
\begin{tabular}{@{}llcccccccc@{}}
\toprule
Recipe & Search & Traj.\ (tasks)/iter & Init & Evolved & $\Delta$ & $\Delta_H$ & $\Delta_M$ & Accepted, $\Sigma\Delta_M$ & GPU-h/iter \\
\midrule
All-evolve            & Seq., fixed-50    & 90--100 (34--43) & 77 & 82 & $+5$ & $+1$ & $+4$ & 1/5, $+4$ & \costFixed{} \\
All-evolve $+$ retry  & Seq., rotating-50 & 86--220 (36--89) & 77 & 84 & $+7$ & $+1$ & $+6$ & 2/6, $+6$ & \costRotating{} \\
\bottomrule
\end{tabular}
\end{table}

\subsection{Corpus manifests}
\Cref{tab:manifests} gives the full manifests summarized in \Cref{tab:recipe};
the 4B winner-only chain of \Cref{tab:main} kept 2, 3 and 2 matched
trajectories in its three iterations, below the top-up floor, and is
described in \Cref{app:scales}.

\begin{table}[h]
\centering
\caption{Corpus manifests behind \Cref{tab:recipe}: trajectories, unique
tasks and prompt/completion pairs offered to the trainer, first to last
iteration of each chain.}
\label{tab:manifests}
\footnotesize\setlength{\tabcolsep}{1.4pt}\renewcommand{\arraystretch}{1.08}
\begin{tabular}{@{}llccccccc@{}}
\toprule
Recipe & Search, evolve set & Traj. & Tasks & Pairs & \shortstack{Per-task\\ cap}
  & Matched & \shortstack{Accepted\\ updates} & $\Sigma\Delta_M$ \\
\midrule
\multicolumn{9}{@{}l}{\textit{Qwen3.5-9B}} \\
All-evolve            & Seq., fixed-50    & 90\,--\,100  & 34\,--\,43 & 361\,--\,532   & 3 & $\times$ & 1 / 5 & $+4$ \\
All-evolve $+$ retry  & Seq., rotating-50 & 86\,--\,220  & 36\,--\,89 & 392\,--\,1189  & 3 & $\times$ & 2 / 6 & $+6$ \\
Mixed siblings        & Tourn.\ $5{+}5$ & 149\,--\,308 & 38\,--\,88 & 618\,--\,1335  & 8 & $\times$ & \textbf{0 / 3} & \textbf{0} \\
\rowcolor{blue!6}
\methodname{}-SFT & Tourn.\ $5{+}5$ & 30\,--\,50 & 30\,--\,50 & 526\,--\,948 & 1\,--\,2 & \checkmark & 2 / 5 & $\mathbf{+6}$ \\
\midrule
\multicolumn{9}{@{}l}{\textit{Qwen3.5-4B}} \\
Mixed siblings          & Tourn.\ $5{+}5$ & 125\,--\,130 & 34\,--\,56 & 1954\,--\,2394 & 8 & $\times$ & 0 / 2 & 0 \\
Winner-only, 1 iter.\ & Tourn.\ $5{+}5$ & 85 & 85         & 1668           & 1 & \checkmark & 0 / 1 & 0 \\
\bottomrule
\end{tabular}
\end{table}

\subsection{Paired records of the reinforcement line}
Every promotion decision logs the per-task win/loss record $(w,\ell)$ against the
incumbent's evaluation. \Cref{tab:flips} lists the six stages of the
reinforcement line for which a paired report was recorded, next to the net
change between the stage scores plotted in \Cref{fig:trajectories}(b). The
two columns do not come from one set of per-task predictions: the paired comparison
evaluates a candidate against the incumbent's own promotion evaluation, whereas the
plotted stage scores come from separate evaluation executions, so they are
shown together only to document measurement variation; the one-to-two-task
differences between the columns are the single-evaluation noise discussed in
\Cref{app:noise}. The flip counts also show how
much churn sits under a small net change (fourteen tasks flipped for a net of
zero at the first harness stage, twenty-three for a net of four at the second
reinforcement stage). \Cref{fig:trajectories}(b) prints only the two records
that reconcile.

\begin{table}[h]
\centering
\caption{Paired per-task records of the reinforcement line, where recorded.
\emph{Paired comparison} columns are from the promotion evaluation; the last
column is the difference of the stage scores of \Cref{fig:trajectories}(b),
which come from separate evaluation executions.}
\label{tab:flips}
\small
\begin{tabular}{@{}lccc@{}}
\toprule
Stage & Paired comparison: wins / losses & Paired comparison: net & Plotted stage scores: $\Delta$ pass count \\
\midrule
Harness 1 & \flipHone{}   & $0$  & $0$  \\
RL 1      & \flipRLone{}  & $0$  & $+1$ \\
RL 2      & \flipRLtwo{}  & $+5$ & $+4$ \\
RL 3      & \flipRLthree{} & $+2$ & $+2$ \\
Harness 4 & \flipHfour{}  & $+2$ & $+2$ \\
RL 4      & \flipRLfour{} & $-3$ & $-5$ \\
\bottomrule
\end{tabular}
\end{table}

\subsection{SWE-bench Lite under both harnesses}
\label{app:swe}
\newcommand{\sweMcnemarP}{\CONFIRMED{0.52}}
\newcommand{\sweOurMcnemarBR}{0.16}
\newcommand{\sweOurMcnemarBS}{0.35}
\newcommand{\sweOurMcnemarSR}{0.02}
\Cref{tab:swe} gives the counts behind \Cref{fig:rq3}(b). Every arm was
graded with the official SWE-bench harness. Under mini-swe-agent the three
checkpoints share one configuration (\Cref{app:prompts}); under the co-evolved
pairing each checkpoint runs inside the HarnessX runner with the harness its
own chain adopted, on the official per-instance images with the task
repository at \texttt{/testbed}, and the diff left in the working tree is the
prediction. The two evolved processors that do not carry over untouched are
marked: the step-budget verifier's budget was moved from 120 to the
benchmark's 250-step cap, and the HTTP verifier-dependency guard evolved for
Tmax server tasks does not fire on SWE-bench. Residual grading errors (two for
base, two for the supervised arm, one for reinforcement; patch-apply and
hanging-test edge cases) are counted as unresolved, so the co-evolved rates
are floors by at most one point.

\paragraph{Paired-instance analysis.} Under mini-swe-agent the
reinforcement checkpoint's margin over base is \sweRLGain{} instances:
\sweBoth{} solved by both, \sweBaseOnly{} by base alone, \sweRLOnly{} by
reinforcement alone, exact McNemar $p=\sweMcnemarP{}$; under the co-evolved
pairs the reinforcement pair resolves \sweOurRLOnlyB{} instances the base
pair does not and loses \sweOurBaseOnlyR{} (\sweOurBothBR{} solved by both,
exact McNemar $p=\sweOurMcnemarBR{}$), the supervised pair wins
\sweOurSFTOnlyB{} and loses \sweOurBaseOnlyS{} against base
($p=\sweOurMcnemarBS{}$), and the reinforcement pair wins \sweOurRLOnlyS{}
and loses \sweOurSFTOnlyR{} against the supervised pair
($p=\sweOurMcnemarSR{}$). Conditional
on reaching the verifier the three checkpoints resolve \sweBaseCondPct\%,
\sweSFTCondPct\%\ and \sweRLCondPct\%\ under mini-swe-agent,
indistinguishable. No-patch outcomes fall from \sweBaseNoPatch{} to
\sweOurBaseEmpty{} for base and from \sweRLNoPatch{} to \sweOurRLEmpty{} for
reinforcement. The arm-to-arm analysis treats each co-evolved model--harness
configuration as the evaluated system, matching the paper's pair-level unit
of analysis. The HarnessX configuration also retains its standard edit tool
after seven steps, while the two Tmax-specific processors described above
remain inactive when their triggers are absent. The mechanisms that transfer
are consistent with this protocol: the edits retained by promotion are loop
and repeated-command breakers (\accLoop\%), budget and
self-verification guards (\accBudget\%) and dependency guards (\accDep\%)
rather than task-specific instructions (\Cref{app:converge}).

\begin{table}[h]
\centering
\caption{SWE-bench Lite (\NsweLite{} instances) for the 9B checkpoints of
\Cref{tab:main} under mini-swe-agent and under the harness each chain adopted.
\emph{No patch} counts instances for which no diff reached the grader. The
finish-reason histogram of the co-evolved runs is no-tool-calls / error /
budget-exceeded, the counterpart of mini-swe-agent's format-error and
limits-exceeded split.}
\label{tab:swe}
\scriptsize
\setlength{\tabcolsep}{3.2pt}\renewcommand{\arraystretch}{1.08}
\begin{tabular}{@{}lcccc|cccc@{}}
\toprule
& \multicolumn{4}{c|}{mini-swe-agent (one harness for all)} & \multicolumn{4}{c}{co-evolved (model, harness) pair} \\
Checkpoint & resolved & unresolved & no patch & format fail. & resolved & unresolved & no patch & finish reasons\textsuperscript{$\dagger$} \\
\midrule
Base   & \sweBaseRes{} (\sweBaseResPct\%) & 88  & \sweBaseNoPatch{} & \sweBaseFormat{} & \sweOurBaseRes{} (\sweOurBaseResPct\%) & \sweOurBaseUnres{} & \sweOurBaseEmpty{} & \sweOurBaseFinish{} \\
$+$SFT & \sweSFTRes{} (\sweSFTResPct\%)   & 71  & \sweSFTNoPatch{}  & \sweSFTFormat{}  & \sweOurSFTRes{} (\sweOurSFTResPct\%)   & \sweOurSFTUnres{}  & \sweOurSFTEmpty{}  & \sweOurSFTFinish{} \\
$+$RL  & \sweRLRes{} (\sweRLResPct\%)     & 106 & \sweRLNoPatch{}   & \sweRLFormat{}   & \textbf{\sweOurRLRes{}} (\textbf{\sweOurRLResPct\%}) & \sweOurRLUnres{} & \sweOurRLEmpty{} & \sweOurRLFinish{} \\
\bottomrule
\multicolumn{9}{@{}l}{\textsuperscript{$\dagger$}\,no tool calls / error / budget exceeded.}
\end{tabular}
\end{table}

\subsection{Chains at other policy scales}
\label{app:scales}
A 4B chain on the mixed-sibling corpus anchors at \fourBanchor{} and never
clears it: its harness candidates score $0$ and $-4$ and its model updates
$-2$ and $-2$, all rejected. A 4B chain on the winner-only corpus anchors at
\fourBnewAnchor{} under the alternative evolve set and stalled. The 4B
winner-only row of \Cref{tab:main} is a three-iteration chain from anchor
\fourBSFTAnchor{}: its harness candidates scored \fourBSFTHarnessSteps{}
against the incumbent and the second was installed, taking the chain to
\fourBSFTFinal{}, while its three model updates scored \fourBSFTModelSteps{}
and were all rejected. Its winner-only corpus kept 2, 3 and 2 trajectory sets
over the three iterations, below the floor that triggers a top-up, so no
iteration trained on a full corpus. Three further completed 4B supervised
chains agree that the model channel does not move at this scale: a
one-iteration winner-only chain from anchor 67 installed a $+1$ harness and
rejected its model update, a three-iteration winner-only chain from anchor 67
rejected every harness and model candidate and ended at 67, and a
three-iteration mixed-siblings chain from anchor 64 ended at 64
(\Cref{tab:ledger}). The 4B reinforcement chain (\Cref{tab:main}) ran three alternating
iterations from anchor \fourBRLAnchor{}: its first reinforcement stage was
promoted from step 50 of a 160-step run after the planned
\fourBRLEpisodesOne{}-episode budget proved too slow, and iterations 2 and 3
ran at \fourBRLEpisodesLate{} episodes, 20 steps and a 16K response budget.
Its harness candidates tied (74) and regressed (68); its reinforcement stages
scored \fourBRLIone{}, \fourBRLItwo{} and \fourBRLIthree{}. The 4B harness-only
control froze the weights for five evolve rounds and installed round 3 at
\fourBHoFinal{} against an anchor of \fourBHoAnchor{}; the per-round scores on
the promotion split were \fourBHoRounds{}. A 2B chain anchors at
\twoBanchor{} with essentially no successful trajectories, marking the
data-starved end of the operating range. A 27B chain anchors at
\twentySevenAnchor{}/\NholdoutTmax{}, marking the ceiling regime on this split.
Together, these endpoints show that the model channel is most productive for a policy
that already solves enough of the curriculum to yield a diverse success corpus
(\Cref{sec:rq:scale}).

\subsection{Round-level traces and the promotion ablation}
\label{app:traces}
\begin{figure}[!htb]
\centering
\includegraphics[width=\linewidth]{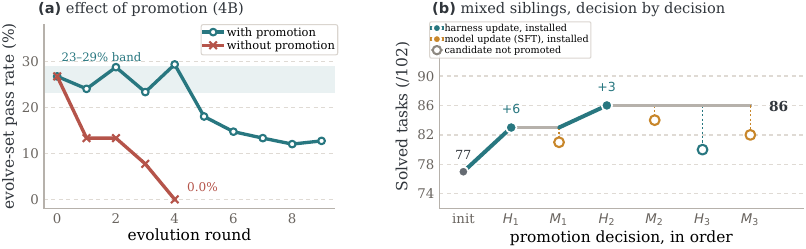}
\caption{\textbf{Promotion ablation and the mixed-siblings trace.} \textbf{(a)}
Without component-wise promotion, a 4B chain falls \ungatedStart\%\ $\rightarrow$ \ungatedEnd\%\ on its
evolve set in five rounds; with promotion, it holds a \gatedBand\%\ band. \textbf{(b)}
The mixed-siblings chain, one promotion decision at a time.}
\label{fig:traces}
\end{figure}

\Cref{fig:traces} collects two views of the chains that the main text
summarizes. Scored before promotion, none of the three 9B regimes of
\Cref{fig:trajectories}(a) is monotone; what separates them is what survives
promotion. Panel (a) isolates the promotion rule on a 4B chain: without promotion,
five rounds of harness evolution fall \ungatedStart\%\ $\rightarrow$
\ungatedEnd\%\ on the evolve set, whereas the promotion-controlled chain holds a
\gatedBand\%\ band. Without component-wise promotion, evolution degrades a working scaffold;
moreover, none of the \nChangesets{} candidate changesets that added a tool was
accepted (\Cref{app:converge}). Panel (b) opens the
mixed-siblings chain one promotion decision at a time: the harness wins twice
($+6$, $+3$) and every model update is rejected.

\paragraph{Promotion retains improvements from a noisy candidate stream.} Stage~B scores every candidate
harness against the incumbent model, so the pure harness effect is identified
and can be pooled over every candidate any chain proposed. Across \dHn{}
candidates from \dHchains{} chains, each of which had already won its
evolve-set search, only \dHpass{} improved the Tmax promotion split, with a
mean effect of \dHmean{} tasks. Harness search therefore produces a noisy
candidate stream, while the deterministic promotion rule allows the chain to retain
improvements and reject regressions. The search also has a floor set by
the policy: five evolve rounds on a frozen 4B score \fourBHoRounds{} of
\NholdoutTmax{} against an anchor of \fourBHoAnchor{}, and the 4B
reinforcement chain rejected every harness candidate it proposed, while the 4B
supervised chain of \Cref{tab:main} installed a \fourBSFTDH{} harness at its
third iteration. Across these runs, policy capability determines which failure
patterns become actionable for harness improvement.

\paragraph{A hypothesis about the order of the channels.} In both
\methodname{} chains (\Cref{fig:trajectories}b, \Cref{fig:rq1}a) the first
productive harness search follows a model update. Our hypothesis is that early failures are largely
procedural, that the first model update internalizes the execution patterns
that fix them, and that the search then faces a smaller, more structured
residual: the late harness win in the reinforcement line flips three tasks,
whereas the first candidate flipped fourteen with zero net gain
(\Cref{tab:flips}). The completed cross-evaluation in \Cref{tab:cross}
provides a direct additivity test, and the same design extends to later
transitions by evaluating the new model under the previous harness.

\subsection{Details behind the analysis}
\label{app:cross}

\paragraph{Reading the cells of \Cref{tab:cross}.} In the \methodname{}-SFT chain,
the model update came first, worth $+4$ under the stock harness, and the
search then found a harness worth $+4$ under the updated policy. The fourth
cell, the base policy under that harness, was evaluated separately and scores
\xcellOne{}: the harness is worth $+4$ under either policy, $I=0$, so for
this transition the model and harness gains are additive and the harness
found after the model update did not require it.

\paragraph{Curriculum drift in numbers.} By the fourth iteration of one chain
\driftEvolveCarried{} of \evolveSetSize{} evolve tasks were carried-over
failures, and the reinforcement chain's third iteration retired \rlRetired{}
of \evolveSetSize{} at once; the cumulative success-filtered corpus spanned
\driftCorpusTasks{} unique tasks at the same point, \driftCorpusCurrent{} from
the current evolve set. In a second chain the evolve-set score fell
\divTwelveEv{} against a promotion score of \divTwelveHo{}
(\Cref{fig:rq1}b). The \methodname{}-SFT chain's late harness candidates lose
$-4$, $-6$ and $-4$ (\Cref{tab:ledger}), and the reinforcement chain's final
candidate, trained after that retirement, is its only rejected reinforcement
update (\rlFinalDrop{} against the incumbent \rlPeak{}). Once the curriculum
has run ahead of the corpus, rollouts land on the hard residual, successes
become scarce, a success-filtered corpus either empties or re-samples history,
and late harness rounds re-propose the same few processor families
(\Cref{app:converge}).

\paragraph{Model updates by mixture.} Every model update trained on the
mixed-siblings corpus scored below the incumbent, \mixDeltas{} tasks at 9B and
$-2$ and $-2$ at 4B, while the matched corpus produced the only supervised
updates above zero under tournament search at 9B ($+4$ and $+2$). The
\methodname{}-SFT chain at 4B rejected its three updates at
\fourBSFTModelSteps{}. The one pilot that holds the trajectory budget fixed
while varying selection is the routing study of \Cref{app:displaced}. A 2B
policy anchors at \twoBanchor{} with almost nothing to harvest, while a 27B
policy at \twentySevenAnchor{} occupies the ceiling regime
(\Cref{app:scales}).

\subsection{Accepted and rejected harness edits}
\label{app:converge}

Across all chains the meta-agent produced \nChangesets{} candidate changesets.
A changeset may touch more than one lever.

\begin{table}[h]
\centering
\caption{Left: what was \emph{proposed} across \nChangesets{} changesets.
Right: what the promotion rule actually \emph{kept}, over
\nAcceptedSlots{} extra processor slots across \nIncumbents{} accepted
incumbents.}
\label{tab:levers}
\small
\begin{minipage}[t]{0.5\textwidth}\centering
\begin{tabular}{@{}lr@{}}
\toprule
Proposed lever & Share \\
\midrule
Add a processor (control)   & \propAddProc\% \\
\;\;\emph{of which} loop / repeat breaker & \addLoop\% \\
\;\;\emph{of which} verifier dependency   & \addDep\% \\
\;\;\emph{of which} step budget / self-verify & \addBudget\% \\
\;\;\emph{of which} other                 & \addOther\% \\
Prompt rewrite (instruction) & \propPrompt\% \\
Remove a processor           & \propRemove\% \\
Empty / no-op copy           & \propNoop\% \\
\textbf{New tool (action)}   & \textbf{\propNewTool\%} \\
\bottomrule
\end{tabular}
\end{minipage}\hfill
\begin{minipage}[t]{0.46\textwidth}\centering
\begin{tabular}{@{}lr@{}}
\toprule
Accepted extras & Share \\
\midrule
Loop / repeat breaker         & \accLoop\% \\
Step budget / self-verify     & \accBudget\% \\
Verifier dependency guard     & \accDep\% \\
\midrule
Prompt rewrite among accepted $H$ & \accPromptRewrite/\nIncumbents{} \\
New tools                     & \textbf{\propNewTool\%} \\
\bottomrule
\end{tabular}
\end{minipage}
\end{table}

Three observations emerge. \textbf{The action lever is never used successfully}:
no accepted harness added a tool, and no such proposal survived promotion. The
search operates almost entirely on control processors and, less often, on the
prompt. \textbf{What survives is narrower than what is proposed}: loop and
repeated-command breakers are \addLoop\%\ of proposed processor additions and
\accLoop\%\ of accepted ones. Finally, \textbf{the two highest-scoring harnesses
both retain the stock five-line prompt}, reaching \frontHonly{} on control
processors alone, in each case a loop breaker plus a guard that installs a
Python dependency the verifier needs. Late rounds converge on variants of the
same two or three mechanism families, consistent with the saturation in
\Cref{sec:rq:drift} and with the argument in \Cref{sec:rq:relay} that model
updates expose new harness-repairable failures. This convergence identifies a
natural trigger for refreshing the search space after a model update.

\section{The Reinforcement Stage}
\label{app:rl}

\subsection{Optimizer and departures from Tmax}
The reinforcement stage runs the open-instruct trainer that Tmax released
\citep{tmax2026}. Each optimizer step samples several rollouts per prompt
under the adopted harness, scores them with the binary verifier, and forms
centered group-relative advantages; groups with zero reward variance carry no
relative signal and are dropped, with the batch refilled by active sampling up
to \rlGroupCap{} prompt groups, the dynamic-sampling device of DAPO
\citep{dapo2025}. The policy update is the DPPO objective \citep{dppo2026}
rather than GRPO-style ratio clipping: the importance ratio is anchored on the
rollout policy's own log-probabilities from the inference engine, and a
per-token trust region masks any update whose binary total-variation
divergence from that policy exceeds $\delta{=}\rlTV{}$ and would move further
away, while updates that move back are never masked. Truncated importance
sampling is off, the LM head is kept in FP32, decoding temperature is 1.0 and
the learning rate is constant. This is Tmax's final recipe for Qwen3.5, which
it adopted over vanilla GRPO for stability on long-horizon terminal rollouts,
and it is what every reported reinforcement stage in this paper ran. Our stage
keeps that algorithmic
core and changes the regime around it (\Cref{tab:rlcompare}): eight rollouts
per prompt rather than 32, to keep unique-task coverage under a budget of
\rlEpisodes{} episodes per stage; two-step rather than four-step asynchrony,
which is the more conservative choice; a small reference-KL coefficient
$\beta{=}\rlBeta{}$ where Tmax uses none, to regularize updates from a few
hundred episodes before the checkpoint re-enters harness evolution; a 16K
rather than 65K response budget; and, the change that is the point of the
experiment, rollouts collected online under the currently adopted harness and
from the evolving task frontier rather than under a fixed harness over a
15k-task pool. The 4B chain used two unique prompts and sixteen samples per
prompt. These settings define the compute-efficient regime evaluated here;
Tmax's reported $G{=}32$ configuration and zero-KL setting provide the
large-budget reference in \Cref{tab:rlcompare}.

\begin{table}[h]
\centering
\caption{Our reinforcement stage against Tmax's published Qwen3.5-9B recipe
\citep{tmax2026}. The algorithmic core is shared; the regime differs.}
\label{tab:rlcompare}
\footnotesize\setlength{\tabcolsep}{4pt}
\begin{tabular}{@{}lll@{}}
\toprule
& This paper & Tmax \\
\midrule
Reward, advantages, objective & outcome-only, centered group-relative, DPPO & same \\
Trust region & binary TV, $\delta{=}\rlTV{}$, rollout log-probs & same \\
Zero-variance groups & dropped; active sampling, $\le$\rlGroupCap{} groups & same \\
LM head / learning rate & FP32 / \rlLR{}, constant & same \\
Samples / unique prompts & \rlSamples{} / \rlPrompts{} (4B: 16 / 2) & 32 / 8 \\
Async steps & \rlAsync{} & 4 \\
Reference KL $\beta$ & \rlBeta{} & 0 \\
Response budget / agent steps & \rlRespLen{} / \rlMaxSteps{} & 65{,}536 / 64 \\
Budget per stage & \rlEpisodes{} episodes, \rlTasks{} tasks & 500 steps $\times$ 256 \\
Task source & frontier (\rlFrontierPct\%) $+$ domain-balanced pool & fixed 15k-task pool \\
Harness during RL & the currently adopted $H^*$ & fixed \\
\bottomrule
\end{tabular}
\end{table}

\subsection{Reward-path debugging record}
Outcome-only RL on this task family initially exhibited a data-path failure
that appeared to be a capability limit. The submission rate remained at
\rlSubmitRateBefore{} for every run: the policy
never emitted the terminal submission action, so every episode scored zero and
every group-relative advantage was zero. Token budget was not the cause;
raising the response budget pushed the truncation rate from \rlTruncBefore{} to
\rlTruncAfter{} without producing a single submission. The cause was in the
data path. The environment's own instance rendering was discarded on reset, the
user turn carried raw taxonomy text instead of the task template, and a
system-prompt override replaced the one remaining turn that mentioned the
submission action; the policy was never told how to submit. Once the prompt
schema was rebuilt to include the instance template, and the sandbox working
directory was pointed at the directory the task images actually populate,
reward became non-zero. We report this because it is the same failure mode the
paper is about: a data-plane defect that can appear as a model or harness
capability bottleneck.

\subsection{Outcome and settings of the reinforcement line}
Alternating reinforcement and harness updates moved the pair
\rlAnchorScore{}~$\rightarrow$~\rlPeak{} on the promotion split,
\rlHarnessShare{} of it through the harness and \rlModelShare{} through the
weights, before a further reinforcement round regressed to \rlFinalDrop{}
(\Cref{fig:trajectories}b, \Cref{tab:ledger}). The run was interrupted by a
cluster outage after its third reinforcement stage and resumed from the saved
incumbent for the fourth iteration. Paired per-task records
were kept for six of the eight stages (\Cref{tab:flips}).

\begin{table}[h]
\centering
\caption{Reinforcement stage settings of the 9B line. The task plane is drawn
from the taxonomy and is disjoint from the promotion split; frontier
exploration spends a fixed fraction of each round on tasks the current pair
has not solved, which is the curriculum idea of \Cref{sec:method:refresh}
applied inside the optimizer rather than around it.}
\label{tab:rlhparams}
\small
\begin{tabular}{@{}llll@{}}
\toprule
\multicolumn{2}{@{}l}{\textbf{Rollout}} & \multicolumn{2}{l}{\textbf{Optimization}} \\
Episodes / iteration    & \rlEpisodes{}   & Unique prompts / step & \rlPrompts{} \\
Training tasks          & \rlTasks{}      & Samples / prompt      & \rlSamples{} \\
Response length         & \rlRespLen{}    & Async steps           & \rlAsync{} \\
Per-turn budget         & \rlPerTurn{}    & Sampled prompt groups & $\le$\rlGroupCap{} \\
Max agent steps         & \rlMaxSteps{}   & Zero-std groups       & dropped \citep{dapo2025} \\
Temperature             & 1.0             & Trust region          & binary TV, $\delta{=}\rlTV{}$ \\
\multicolumn{2}{@{}l}{\textbf{Compute}} & Reference KL $\beta$ / LM head & \rlBeta{} / FP32 \\
Frontier exploration    & \rlFrontierPct\% & Truncated IS          & off (rollout log-probs) \\
Learners / vLLM replicas & \rlLearners{} / \rlVllm{} & Wall-clock cap & \rlWall\,h \\
\bottomrule
\end{tabular}
\end{table}

This line uses \rlEpisodes{} episodes on \rlTasks{} tasks on a single node,
compared with a reference recipe using roughly 128k episodes and longer
responses across eight nodes. The resulting gains demonstrate effective
on-policy learning under the evolved harness in a substantially smaller
compute regime.

\section{Additional Pilots}
\label{app:displaced}

\subsection{Routing pilot on Terminal-Bench~2.1}
\Cref{tab:routing} is an early single-benchmark
pilot on Terminal-Bench 2.1 that isolates corpus construction under a fixed
baseline harness and without a promotion step. Holding the trajectory budget
fixed while varying selection provides a complementary controlled comparison
to the complete-recipe study in the main text. It ranks capability-routed selection
(\routeCapability{}) above taxonomy-balanced selection (\routeTaxonomy{}) and
an undifferentiated mix (\routeMix{}) at equal corpus size. The default recipe
of \Cref{tab:recipe} is current-first with history capped at \histCap\%\ and
tops up to \sftGenTarget{} unique tasks at one trajectory each, so it buys
coverage without volume.

\begin{table}[!htb]
\centering
\caption{Routing pilot. Qwen3.5-9B, baseline harness, \routeN{}
Terminal-Bench 2.1 tasks, mean of \routeReps{} runs.}
\label{tab:routing}
\small
\begin{tabular}{@{}lcc@{}}
\toprule
Corpus construction & Trajectories & Solved $\uparrow$ \\
\midrule
Base policy, no SFT                & ---            & \routeBase{} \\
Undifferentiated mix               & 107            & \routeMix{} \\
Balanced by task taxonomy          & \routeCorpus{} & \routeTaxonomy{} \\
\textbf{Routed by capability need} & \routeCorpus{} & \textbf{\routeCapability{}} \\
\bottomrule
\end{tabular}
\end{table}

\subsection{Case study of grounded harness edits}
\Cref{tab:case} follows four consecutive
harness edits from a pilot chain on a different policy family, each grounded in
a cited failure. It illustrates what the meta-agent proposes and what survives;
it is not part of the terminal-task record.

\begin{table}[!htb]
\centering
\caption{Four grounded harness edits on the evolve subset. Every diagnosis was
correct; not every intervention helped. This is why proposal and promotion are
separated, and why promotion is deterministic.}
\label{tab:case}
\small
\begin{tabular}{@{}llll@{}}
\toprule
Round & Observed failure & Harness edit & Effect \\
\midrule
4B R1  & Bash calls written as text, never executed & inline tool-call recovery & \caseFourRone{} \\
31B R2 & Backend rejected the tool transport; all tasks died at step 0 & transport bypass & \caseThirtyRtwo{} \\
31B R3 & Model did not create the files the verifier required & required-output guard & \caseThirtyRthree{} \\
31B R4 & Model assumed unavailable commands existed & environment preflight & \caseThirtyRfour{} \\
\bottomrule
\end{tabular}
\end{table}

\end{document}